\documentclass[letterpaper]{article}

\usepackage[T1]{fontenc}

\usepackage{geometry}
\usepackage{setspace}

\usepackage[
    colorlinks=true,
    linkcolor=black,
    citecolor=black,
    urlcolor=black
]{hyperref}
\usepackage{booktabs}
\usepackage{multirow}
\usepackage{array}
\usepackage{subcaption}
\usepackage{amssymb}
\usepackage[
  style=chem-acs,
  sorting=none
]{biblatex}
\usepackage{graphicx}
\usepackage{float}
\newfloat{scheme}{htbp}{los}
\floatname{scheme}{Scheme}
\floatname{chart}{Chart}
\newfloat{graph}{htbp}{loh}

\usepackage{chemformula} 
\usepackage[version = 4]{mhchem} 
\usepackage{authblk}

\title{HierDTA: Hierarchical Structure-Aware Representation Learning for Drug--Target Affinity Prediction}

\author[1,2]{Huangchao Xu}
\author[1,*]{Baohua Zhang}
\author[1]{Qian Liu}
\author[1]{Zhong Jin}
\author[1]{Yi Du}
\affil[1]{Computer Network Information Center, Chinese Academy of Sciences, Beijing 100083, China}
\affil[2]{School of Computer Science and Technology, University of Chinese Academy of Sciences, Beijing 101408, China}

\date{*Email: zhangbh@cnic.cn}

\begin{document}

\maketitle

\begin{abstract}
  Drug--target affinity (DTA) prediction plays an important role in computer-aided drug discovery, where reliable computational models can help prioritize compound--protein pairs before experimental validation. A key challenge in DTA prediction is to effectively mine multi-level molecular information from drug structures while accurately modeling the binding-site environment of target proteins. To address these issues, we introduce HierDTA, a hierarchical structure-aware framework for DTA prediction. \textcolor{black}{HierDTA represents each drug as an atom--motif heterogeneous graph and introduces an augmentation module that propagates functional-motif semantics to the corresponding atoms. For proteins, kinase-domain structures are converted into residue interaction graphs. The resulting atom and residue representations are integrated using cross-attention to capture pair-specific relationships, while molecule fingerprints and pretrained ESM-2 embeddings provide complementary global information. Compared with state-of-the-art methods, HierDTA achieves superior performance on the Davis dataset across all metrics and attains the highest \(r_m^2\) of 0.791 on KIBA. It also exhibits strong generalization under cold-start settings, particularly on KIBA, where MSE is reduced by 21.7\% in the unseen-protein scenario. Ablation study and interpretability analysis supported the contributions of the hierarchical augmentation and interaction components.} Case studies further showed that HierDTA provided early enrichment of candidate bioactive compounds. These results indicate that HierDTA is an effective framework for affinity prediction and target-specific virtual screening.

\end{abstract}





\section{Introduction}

Accurate prediction of drug--target affinity (DTA) plays a fundamental role in computer-aided drug discovery by estimating the binding strength between candidate compounds and biological targets. Reliable affinity prediction enables the prioritization of promising drug candidates before costly experimental validation, thereby reducing both the time and expense associated with drug development. Traditional computational approaches, including molecular docking and molecular dynamics simulations, provide physically meaningful estimates of binding affinity but remain computationally prohibitive for large-scale virtual screening. With the rapid accumulation of biochemical data, deep learning (DL) has emerged as an efficient alternative, learning complex interaction patterns directly from molecular data while achieving remarkable predictive performance.

The evolution of DL-based DTA prediction has been closely accompanied by the continuous advancement of molecular representations, progressing from sequence-based to graph-based, three-dimensional structural modeling, and multimodal representation learning. Early methods primarily represented drugs and proteins as one-dimensional (1D) sequences. Representative models such as DeepDTA~\cite{10.1093/bioinformatics/bty593} employed convolutional neural networks (CNNs) to encode drug SMILES strings and protein amino acid sequences, demonstrating that deep neural networks could effectively learn affinity-related features without handcrafted descriptors. However, sequence-based methods inevitably discard structural information that is crucial for molecular recognition and binding. To better preserve molecular topology, subsequent studies introduced graph-based molecular representations. GraphDTA~\cite{10.1093/bioinformatics/btaa921} first represented drug molecules as molecular graphs and employed graph neural networks (GNNs) for molecular feature learning. Building upon this paradigm, DGraphDTA~\cite{D0RA02297G} incorporated protein residue contact graphs, while MGraphDTA~\cite{D1SC05180F} explored multi-view graph fusion to exploit complementary structural information. \textcolor{black}{Attention-based methods~\cite{10.1109/TCBB.2022.3170365, yuan2022fusiondta, WU2024623} }further enhanced graph representations through attention mechanisms and more effective feature aggregation.

Recent advances in protein structure prediction, particularly AlphaFold~\cite{jumper2021highly}, have shifted DTA prediction toward structure-aware modeling. Instead of representing entire proteins, emerging methods increasingly focus on protein binding \textcolor{black}{regions} that directly mediate molecular recognition. Representative approaches such as 3DProtDTA~\cite{D3RA00281K} and PocketDTA~\cite{zhao2024pocketdta} construct protein graphs from binding-region structures and employ graph neural networks to capture spatial interactions among residues. Meanwhile, drug representations have also evolved beyond two-dimensional molecular topology to incorporate three-dimensional conformation.

More recently, multimodal and multiscale learning has emerged as a promising direction for improving
DTA prediction by integrating complementary biochemical information from multiple modalities. Rather
than relying on a single representation, recent methods jointly exploit sequence, graph, and structural information
to obtain more comprehensive molecular embeddings. NHGNN-DTA~\cite{he2023nhgnn}, MMSG-DTA~\cite{xu2025mmsg}, and DMFF-DTA~\cite{he2025dual} combine heterogeneous molecular representations through hybrid or multiscale fusion strategies. Beyond feature fusion, several methods explicitly interactions within and between drugs and targets. MSGNN-DTA~\cite{ijms24098326} uses mutual attention, GRA-DTA~\cite{ijms25105126} applies relation-aware attention, and \textcolor{black}{MT-DTA~\cite{zhu2023associative} combines CNN--Transformer encoders with associative learning and autoencoder-based reconstruction to capture intramolecular dependencies and intermolecular relationships. Related efforts have incorporated diffusion-based graph operations to improve multiscale structural modeling. TDGraphDTA~\cite{zhu2023tdgraphdta} and MDCT-DTA~\cite{zhu2024mdctdta} broaden the receptive field of molecular graphs and integrate drug and target features at multiple levels through Transformer-based or interlayer interaction mechanisms. RRGDTA~\cite{zhu2025rrgdta} further introduces rotary encoding and information-retention modules to strengthen structural correlations while preserving local and global association information. More recent architectures have explored alternative fusion and interaction strategies}. KANPM-DTA~\cite{rakib2026kanpm} combines ESMC-guided protein graphs with gated feature fusion and a Kolmogorov--Arnold Network predictor, whereas PMHGT-DTA~\cite{zhang2025multimodal} uses hierarchical graph transformers and cross-attention to characterize local and global structural patterns and atom--residue relationships. \textcolor{black}{Related biomedical studies further support this direction: meta-relation-guided heterogeneous graph transformers can capture complementary local and global associations among diverse biomolecules~\cite{guo2026circrna}, while diffusion-based augmentation can learn biologically plausible protein features under sparse supervision~\cite{liu2026diffad}. These advances highlight the value of preserving heterogeneous relationships and biologically meaningful structure during feature learning.} 

\textcolor{black}{Nevertheless, existing approaches generally derive multiscale chemical features from receptive fields or latent representations. Direct information exchange between atomic topology and chemically defined functional motifs has received limited attention. In addition, although structure-based protein modeling has improved the characterization of targets, many residue graphs are constructed primarily from spatial proximity and do not explicitly distinguish the physicochemical interactions among residues.}


To address these issues, we propose HierDTA, a structure-aware framework for DTA prediction. \textcolor{black}{On the drug side, motif-level information is explicitly propagated to atom representations to capture hierarchical molecular features. On the protein side, physicochemical edge attributes are incorporated into residue interaction graphs to describe residue-level structural relationships. HierDTA then combines local atom--residue interactions with global drug and protein representations for affinity prediction}. The main contributions of this work are summarized as follows:



    

\begin{itemize}
\item \textcolor{black}{We introduce a motif-to-atom augmentation module that explicitly propagates chemical semantics from functional motifs to their constituent atoms. This top-down information pathway enriches atom-level representations while preserving molecular topology.}

\item \textcolor{black}{We construct kinase-domain residue interaction graphs and develop an edge-aware graph encoder that incorporates physicochemical attributes during residue-level message passing, enabling structurally informed protein representation learning.}

\item \textcolor{black}{We employ cross attention between drug atoms and protein residues, offering fine-grained modeling of drug-target interactions. These components are unified within HierDTA, with attention-based visualizations revealing motif semantic propagation and key interactions. Experiments on Davis and KIBA under both warm and cold-start settings, together with case studies, demonstrate the model's effectiveness and practical applicability.}
\end{itemize}

\begin{figure*}[htbp]
    \centering
    \includegraphics[width=\textwidth]{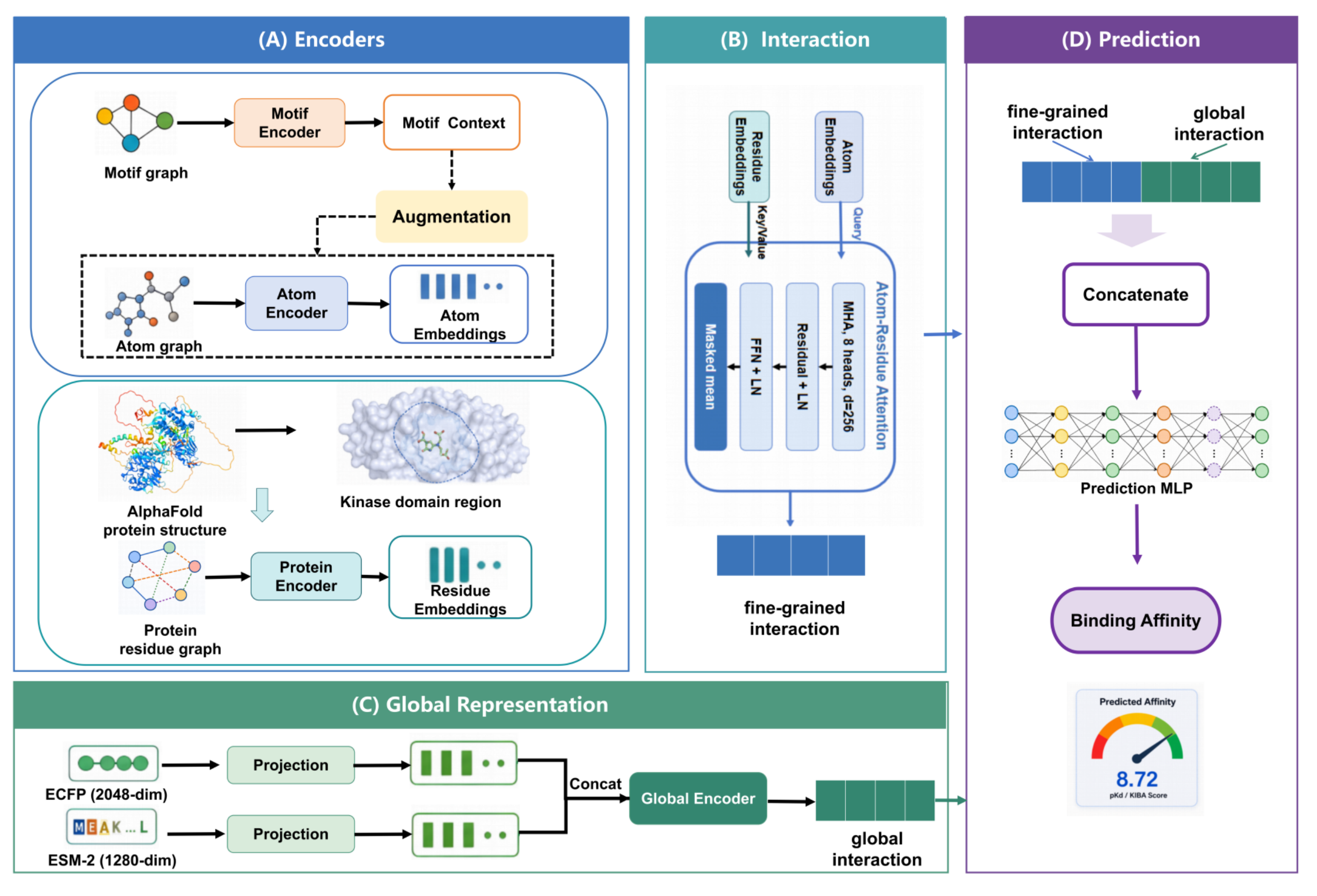}
    \caption{\textcolor{black}{Architecture of the proposed framework.(A) Encoders. Drug features are encoded from motif and atom graphs, with motif-derived context augmenting atom representations, while protein residue features are extracted from the kinase-domain structure. (B) Interaction. Atom and residue tokens are integrated through multi-head attention to capture fine-grained drug–protein interactions. (C) Global Representation. ECFP fingerprints and ESM-2 embeddings are projected and fused to obtain a global representation. (D) Prediction. Local and global interaction features are concatenated and fed into an MLP to predict binding affinity.}}
    \label{fig:architecture}
\end{figure*}

\section{Methods}

\subsection{Drug Representation}
HierDTA adopts a hierarchical molecular representation that jointly models atomic structures and functional motifs. Given a SMILES string, a molecular graph is first constructed using RDKit~\cite{landrum2013rdkit}, where atoms and covalent bonds are represented as graph nodes and edges, respectively. Functional motifs are subsequently identified through a rule-based functional group decomposition algorithm~\cite{han2023himgnn}, which detects chemically meaningful functional groups and ring structures. Based on these components, a heterogeneous molecular graph consisting of atom nodes and motif nodes is established. \textcolor{black}{Formally, the molecular graph is denoted as $\mathcal{G}_d=(\mathcal{V}_a,\mathcal{V}_m,\mathcal{E}_{aa},\mathcal{E}_{am},\mathcal{E}_{mm})$. Here, $\mathcal{V}_a$ and $\mathcal{V}_m$ represent the atom and motif node sets, respectively. $\mathcal{E}_{aa}$, $\mathcal{E}_{am}$ and $\mathcal{E}_{mm}$ denote atom--atom, atom--motif, motif--motif edge sets. The corresponding node and edge feature matrices are represented by $\mathbf{X}_a$, $\mathbf{X}_m$, $\mathbf{E}_{aa}$, $\mathbf{E}_{am}$, and $\mathbf{E}_{mm}$.} \textcolor{black}{The complete motif decomposition procedure is provided in the Supporting Information S1.}

\textcolor{black}{Each atom--motif edge connects an atom to every motif that contains it and is assigned a membership indicator with a value of 1. These membership relations establish the explicit correspondence between the two molecular levels and are subsequently used by the augmentation module to route motif-derived chemical context to the associated atoms.} The original molecular topology is preserved through atom--atom edges, while motif--motif edges are constructed between motifs sharing common atoms or connected through inter-motif chemical bonds. Node and edge features are initialized using chemically meaningful descriptors summarized in Table~\ref{tab:drug_features}. Atom nodes are encoded by physicochemical properties including atom type and atomic number, whereas motif nodes aggregate the information of constituent atoms and bonds. Edge features describe chemical relationships between neighboring atoms or neighboring motifs according to their structural connections.

Since graph message passing mainly captures local structural contexts, a 2048-dimensional Extended-Connectivity Fingerprint (ECFP4)~\cite{rogers2010extended} is further incorporated to preserve global molecular information. The hierarchical graph representation and molecular fingerprint therefore provide complementary descriptions of molecules, enabling both local chemical environments and global scaffold information to be exploited during affinity prediction.

\begin{table}[htbp]
\centering
\caption{\textcolor{black}{Feature definition of the hierarchical molecular graph for drug representation.}}
\label{tab:drug_features}
\begin{tabular}{p{13cm}c}
\hline
\textbf{Name} & \textbf{Size} \\
\hline

\textbf{Atom node features} & \textbf{37} \\

\quad atom type (B, Br, C, Ca, Cl, F, H, I, N, Na, O, P, S) & 13 \\
\quad atomic number & 1 \\
\quad explicit valence (0--6) & 7 \\
\quad total hydrogen count (0--4) & 5 \\
\quad atom hybridization (SP, SP$^2$, SP$^3$, SP$^3$D, SP$^3$D$_2$) & 5 \\
\quad is atom aromatic (1--yes, 0--no) & 1 \\
\quad is atom in a ring (1--yes, 0--no) & 1 \\
\quad chirality (R, S, Unknown) & 3 \\
\quad has possible chirality (1--yes, 0--no) & 1 \\

\textbf{Motif node features} & \textbf{50} \\

\quad aggregation of atom features within the motif & 37 \\
\quad aggregation  of internal bond features & 13 \\

\textbf{Atom--Atom edge features} & \textbf{13} \\

\quad bond type (Single, Double, Triple, Aromatic) & 4 \\
\quad is bond in a ring (1--yes, 0--no) & 1 \\
\quad is bond conjugated (1--yes, 0--no) & 1 \\
\quad bond stereochemistry (None, Any, Z, E, Cis, Trans, Unknown) & 7 \\

\textbf{Atom--Motif edge features} & \textbf{1} \\
\quad atom-to-motif membership indicator & 1 \\

\textbf{Motif--Motif edge features} & \textbf{37} \\

\quad aggregation of shared atom features or bonded atom features & 37 \\

\hline
\end{tabular}
\end{table}

\subsection{Protein Representation}


\textcolor{black}{
Residue-level protein graphs were constructed from predicted 3D structures of target proteins obtained from the AlphaFold protein structure database~\cite{jumper2021highly}. Given that both datasets are primarily composed of kinase targets, full-length structures were further restricted to kinase-related domains before graph construction. UniProt domain annotations~\cite{uniprot2019uniprot} were used to extract regions related to kinase activity, including protein kinase, histidine kinase. Thus, the structural boundaries were defined by kinase-related domain annotations. 
This preprocessing step not only decreased the noise in the data but also eliminated most residues with a low per-residue confidence score (pLDDT). Because low-confidence residues frequently occur in flexible or disordered regions outside well-characterized domains, the filtering procedure may also reduce the influence of uncertain structural regions. The pLDDT distribution is reported in the Supporting Information S2.
}


\begin{table}[htbp]
\centering
\caption{\textcolor{black}{Feature definition of the residue interaction graph for protein representation.}}
\label{tab:protein_features}
\begin{tabular}{p{11cm}c}
\hline
\textbf{Name} & \textbf{Size}\\
\hline

\textbf{Residue node features} & \textbf{41}\\

\quad solvent-accessible surface area (normalized) & 1\\
\quad phi angle ($\phi$) & 1\\
\quad psi angle ($\psi$) & 1\\
\quad one-hot encoded secondary structure &
6\\
\quad AAPHY7 physicochemical descriptors &
7\\
\quad BLOSUM62 evolutionary descriptors &
23\\
\quad phosphorylation status &
1\\
\quad mutation status &
1\\

\textbf{Residue--Residue edge features} & \textbf{10}\\

\quad covalent bond (1--yes, 0--no) & 1\\
\quad hydrophobic contact (1--yes, 0--no) & 1\\
\quad hydrogen bond, donor to acceptor (1--yes, 0--no) & 1\\
\quad hydrogen bond, acceptor to donor (1--yes, 0--no) & 1\\
\quad salt bridge, cation to anion (1--yes, 0--no) & 1\\
\quad salt bridge, anion to cation (1--yes, 0--no) & 1\\
\quad cation--$\pi$ interaction, aromatic ring to cation (1--yes, 0--no) & 1\\
\quad cation--$\pi$ interaction, cation to aromatic ring (1--yes, 0--no) & 1\\
\quad parallel $\pi$--$\pi$ stacking (1--yes, 0--no) & 1\\
\quad perpendicular $\pi$--$\pi$ stacking (1--yes, 0--no) & 1\\

\hline
\end{tabular}
\end{table}

The filtered protein structures are converted into residue graphs \textcolor{black}{with physicochemical interaction definitions inspired by the Open Drug Discovery Toolkit (ODDT)~\cite{wojcikowski2015open}}. Each residue graph is denoted as $\mathcal{G}_p=(\mathcal{V}_r,\mathcal{E}_{rr})$, where $\mathcal{V}_r$ denotes the residue node set and $\mathcal{E}_{rr}$ denotes the residue--residue interaction edge set. \textcolor{black}{An edge is established when two residues are covalently connected or satisfy the criteria for a specified noncovalent interaction. The interaction types and their corresponding criteria are summarized in Supporting Information S2.}

Residue nodes are characterized by structural and biochemical descriptors, including solvent-accessible surface area (SASA), backbone torsion angles ($\phi$ and $\psi$), secondary structure, AAPHY7 physicochemical descriptors, BLOSUM62 evolutionary descriptors, phosphorylation status, and mutation status, as summarized in Table~\ref{tab:protein_features}. The residue node feature matrix is denoted as $\mathbf{X}_r$. \textcolor{black}{Each edge in $\mathcal{E}_{rr}$ is associated with a 10-dimensional attribute vector, and the residue--residue edge feature matrix is denoted as $\mathbf{E}_{rr}\in\mathbb{R}^{|\mathcal{E}_{rr}|\times 10}$.
Its channels encode covalent bonds, hydrophobic contacts, hydrogen bonds, salt bridges, cation--$\pi$ interactions, and $\pi$--$\pi$ stacking. Separate directional channels are retained for hydrogen bonds, salt bridges, and cation--$\pi$ interactions to distinguish donor--acceptor, cation--anion, and aromatic--cation roles, respectively. The node and edge feature definitions are summarized in Table~\ref{tab:protein_features}}. To complement structural information, protein sequences are further encoded by the pretrained ESM-2 protein language model~\cite{rives2021biological,lin2023evolutionary}, providing global evolutionary semantics that cannot be fully captured through residue graphs.

\subsection{Model Architecture}
The overall architecture of HierDTA is illustrated in Figure~\ref{fig:architecture}. \textcolor{black}{For each drug--target pair, the drug encoder first learns atom and motif representations from the molecular graph and propagates motif-derived chemical semantics to the corresponding atoms through the augmentation module. In parallel, the protein encoder applies edge-aware message passing to the kinase-domain residue graph to obtain residue representations. The resulting atom and residue embeddings are integrated by cross-attention to derive fine-grained interaction representations. A complementary global representation is constructed separately from the drug ECFP4 fingerprint and the protein ESM-2 embedding. Finally, the fine-grained and global features are concatenated and supplied to the prediction head to estimate binding affinity.}

\subsubsection{Drug Encoder}
\textcolor{black}{The drug encoder operates on the atom and motif subgraphs of $\mathcal{G}_d$ while preserving their correspondence through the atom--motif membership edges $\mathcal{E}_{am}$. The initial atom and motif features are first projected into a shared $d$-dimensional latent space,}

\begin{equation}
\color{black}
\mathbf{H}_a^{(0)}=\phi_a(\mathbf{X}_a),
\qquad
\mathbf{H}_m^{(0)}=\phi_m(\mathbf{X}_m),
\end{equation}

\textcolor{black}{where $\phi_a$ and $\phi_m$ are independently parameterized linear transformations followed by ReLU activation. Atom--atom connectivity and motif--motif organization are then modeled by separate graph encoders~\cite{brody2021attentive}. The motif update is expressed as}

\begin{equation}
\color{black}
\mathbf{H}_m^{(l)}=
\operatorname{GATv2Conv}_m^{(l)}
\left(
\mathbf{H}_m^{(l-1)},
\mathcal{E}_{mm},
\mathbf{E}_{mm}
\right),
\end{equation}

\textcolor{black}{and the intermediate output of the $l$-th atom-encoding layer is obtained analogously,}

\begin{equation}
\color{black}
\widetilde{\mathbf{H}}_a^{(l)}=
\operatorname{GATv2Conv}_a^{(l)}
\left(
\mathbf{H}_a^{(l-1)},
\mathcal{E}_{aa},
\mathbf{E}_{aa}
\right).
\end{equation}

\textcolor{black}{Here, $\mathbf{H}_m$ summarizes the chemical context of functional motifs, while $\widetilde{\mathbf{H}}_a^{(l)}$ captures the local atomic messages encoded at layer $l$. The two encoders have independent parameters, thereby allowing atom- and motif-level connectivity to be modeled at their respective representational scales.}

\begin{figure*}[htbp]
  \centering
  \includegraphics[width=0.7\textwidth]{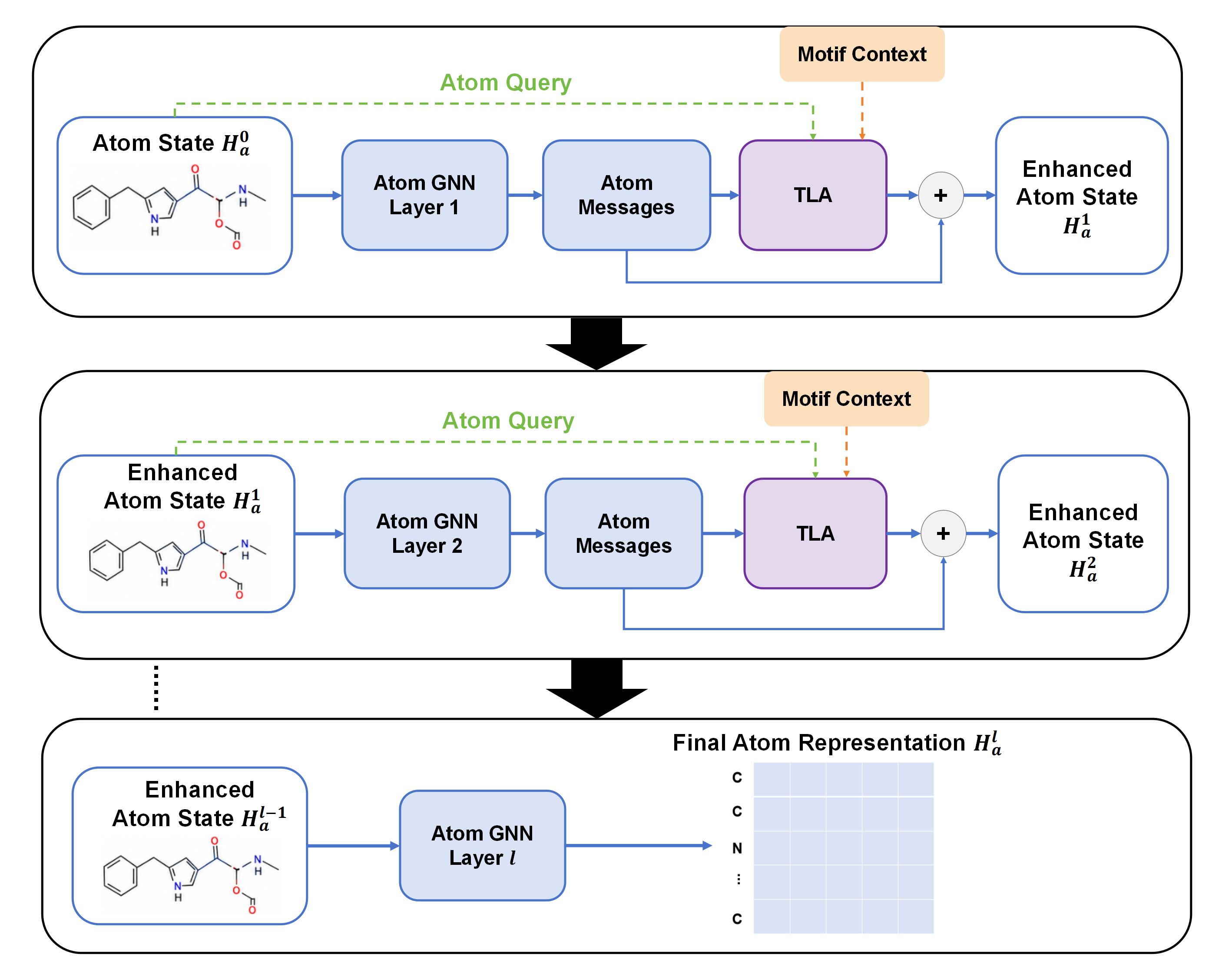}
  \caption{\textcolor{black}{Architecture of the motif-to-atom augmentation module. Motif representations are propagated to their constituent atoms through atom--motif membership edges and aggregated to obtain motif-derived semantic messages. For each atom, TLA operation uses the incoming atom representation as the query and attends to the motif context and atom message. The resulting augmentation message is incorporated into the atom representation through a residual connection before subsequent atom-level message passing.}}
  \label{fig:augmentation}
\end{figure*}

\textcolor{black}{To establish top-down information flow within the hierarchical molecular graph, we propose a motif-to-atom augmentation module, as
illustrated in Figure~\ref{fig:augmentation}. The module transfers the encoded motif semantics to their constituent atoms. For atom $i$, the representations of all motifs connected to that atom through $\mathcal{E}_{am}$ are mean-aggregated to obtain a motif-derived semantic context,}

\begin{equation}
\color{black}
\mathbf{m}_i=
\frac{1}{|\mathcal{N}_m(i)|}
\sum_{j\in\mathcal{N}_m(i)}
\mathbf{h}_{m,j},
\end{equation}


\textcolor{black}{where $\mathcal{N}_m(i)$ denotes the set of motifs containing atom $i$. Mean aggregation permits information from overlapping functional motifs to be integrated without introducing a dependence on the number of motif memberships. We then employ a Transformer-based Local Augmentation (TLA) operation to adaptively integrate the motif-derived semantic context $\mathbf{m}_i$ with the atom-derived structural message $\widetilde{\mathbf{h}}_{a,i}^{(l)}$. Specifically, TLA formulates this local fusion as a transformer attention operation, in which the atom representation entering the current GAT layer serves as the query, while the two complementary messages, $\mathbf{m}_i$ and $\widetilde{\mathbf{h}}_{a,i}^{(l)}$, form the key--value sequence. This design enables each atom to selectively attend to motif-level semantic information and atom-level structural information according to its current representation.}

\begin{equation}
\color{black}
\mathbf{q}_i^{(l)}=\mathbf{h}_{a,i}^{(l-1)},
\qquad
\mathbf{K}_i^{(l)}=\mathbf{V}_i^{(l)}=
\left[
\mathbf{m}_i;
\widetilde{\mathbf{h}}_{a,i}^{(l)}
\right].
\end{equation}

\textcolor{black}{For attention head $k$, the augmentation message is computed as}

\begin{equation}
\color{black}
\mathbf{u}_{i,k}^{(l)}=
\operatorname{softmax}
\left(
\frac{
(\mathbf{q}_i^{(l)}\mathbf{W}_{Q,k})
(\mathbf{K}_i^{(l)}\mathbf{W}_{K,k})^{\mathsf{T}}
}{\sqrt{d_k}}
\right)
\mathbf{V}_i^{(l)}\mathbf{W}_{V,k},
\end{equation}

\textcolor{black}{where $d_k$ is the dimensionality of each attention head and $\mathbf{W}_{Q,k}$, $\mathbf{W}_{K,k}$, and $\mathbf{W}_{V,k}$ are learnable projection weights. The outputs of all heads are concatenated,}

\begin{equation}
\color{black}
\mathbf{u}_i^{(l)}=
\operatorname{Concat}
\left(
\mathbf{u}_{i,1}^{(l)},\ldots,\mathbf{u}_{i,H}^{(l)}
\right),
\end{equation}

\textcolor{black}{and the atom representation is updated through a residual connection,}

\begin{equation}
\color{black}
\mathbf{h}_{a,i}^{(l)}=
\widetilde{\mathbf{h}}_{a,i}^{(l)}+
\mathbf{u}_i^{(l)}.
\end{equation}

\textcolor{black}{The augmentation is inserted between successive atom-encoding layers, such that the injected motif semantics can participate in subsequent atom-level message passing. The final output $\mathbf{H}_a$ therefore comprises motif-enhanced atom representations and is retained at the node level for subsequent atom--residue interaction modeling.}

\subsubsection{Protein Encoder}
Given the residue graph $\mathcal{G}_p=(\mathcal{V}_r,\mathcal{E}_{rr})$ and residue feature $\mathbf{X}_r$, an \textcolor{black}{edge-aware} graph encoder~\cite{xu2018powerful} is adopted to update residue representations,

\begin{equation}
\color{black}
\mathbf{H}_r^{(l+1)}=
\operatorname{GINEConv}^{(l)}
\left(
\mathbf{H}_r^{(l)},
\mathcal{E}_{rr},
\mathbf{E}_{rr}
\right)
.
\end{equation}


\textcolor{black}{The $\operatorname{GINEConv}$ operator incorporates the 10-dimensional edge attribute matrix $\mathbf{E}_{rr}$ directly into residue-level message passing, thereby distinguishing messages associated with different physicochemical interaction types. After $L$ layers, the protein encoder yields $\mathbf{H}_r$, served as residue tokens for atom--residue cross attention in Section~\ref{sec:interaction_module}.}

\subsubsection{Interaction}
\label{sec:interaction_module}
\textcolor{black}{The interaction module captures pair-specific associations between the motif-enhanced atom representations $\mathbf{H}_a$ and the residue representations $\mathbf{H}_r$. Since individual ligand atoms may interact with different protein environments, atom representations are used as queries, while residue representations serve as keys and values. The query, key, and value matrices are defined as}

\begin{equation}
\color{black}
\mathbf{Q}=\mathbf{H}_a\mathbf{W}_Q,
\qquad
\mathbf{K}=\mathbf{H}_r\mathbf{W}_K,
\qquad
\mathbf{V}=\mathbf{H}_r\mathbf{W}_V.
\end{equation}

\textcolor{black}{Multi-head cross-attention is then employed to capture atom--residue dependencies across multiple representation subspaces, yielding fine-grained interaction features}

\begin{equation}
\color{black}
\mathbf{H}_{\mathrm{inter}}
=\operatorname{CrossAttention}(\mathbf{Q},\mathbf{K},\mathbf{V}).
\end{equation}

\textcolor{black}{The resulting interaction features are aggregated by mean pooling to obtain the pair-level representation $\mathbf{z}_{\mathrm{inter}}$. The corresponding attention coefficients are retained for interpretability analysis in Section~\ref{sec:interpret}.}

\subsubsection{Global Representation}

\textcolor{black}{The interaction module characterizes associations between atom and residue nodes, whereas the global branch provides complementary information concerning the overall molecular scaffold and protein sequence. Specifically, the ECFP4 fingerprint and the ESM-2 embedding are independently projected into latent representations,}

\begin{equation}
\color{black}
\mathbf{z}_{fp}=\operatorname{FC}_{1}(\mathbf{fingerprint}),
\qquad
\mathbf{z}_{esm}=\operatorname{FC}_{2}(\mathbf{esm}),
\end{equation}

\begin{equation}
\color{black}
\mathbf{z}_{\mathrm{global}}=
\operatorname{FC}
\left(
[\mathbf{z}_{fp};\mathbf{z}_{esm}]
\right).
\end{equation}

\textcolor{black}{where $\operatorname{FC}_{1}$, $\operatorname{FC}_{2}$ and FC are multilayer perceptions. The global branch is constructed to complement the fine-grained interaction representation with molecule fingerprint and protein pretrained sequence-level semantics.}

\subsubsection{Prediction Head}


\textcolor{black}{The prediction head integrates the atom--residue interaction representation with the global drug--protein representation by concatenation,}

\begin{equation}
\color{black}
\mathbf{z}_{dp}=
[\mathbf{z}_{\mathrm{inter}};\mathbf{z}_{\mathrm{global}}].
\end{equation}

This representation is passed to a multilayer perceptron to generate the predicted affinity score,

\begin{equation}
\hat{y}=f_{\theta}(\mathbf{z}_{dp}),
\end{equation}

where $f_{\theta}(\cdot)$ denotes the MLP predictor and $\hat{y}$ is the predicted affinity. The model is optimized end-to-end using the mean squared error loss,

\begin{equation}
\mathcal{L}
=
\frac{1}{N}
\sum_{i=1}^{N}
\left(
y_i-\hat{y}_i
\right)^2,
\end{equation}

where $N$ is the number of training samples in a batch, and $y_i$ and $\hat{y}_i$ are the ground-truth and predicted affinity values for the $i$-th sample, respectively.

\section{Experiment and Metrics}
\subsection{Datasets}

We evaluate HierDTA on two widely used benchmark datasets for drug--target affinity prediction, namely Davis~\cite{davis2011comprehensive} and KIBA~\cite{tang2014making}. The Davis dataset contains binding affinities between 68 kinase inhibitors and 442 kinase proteins, resulting in 30,056 drug--target interaction pairs. Binding affinities are measured by the dissociation constant ($K_d$) and transformed into the logarithmic scale ($pK_d$) to improve numerical stability during model training.

The KIBA dataset integrates multiple experimentally measured bioactivity indicators, including $K_i$, $K_d$, and $IC_{50}$, into a unified KIBA score. It consists of 118,083 interaction pairs involving 2,111 drugs and 228 target proteins, covering a substantially broader chemical and biological space than the Davis dataset. The larger scale and higher diversity of KIBA provide a more challenging benchmark for evaluating the robustness and generalization capability of DTA prediction models.

\subsection{Evaluation Strategies}
To comprehensively evaluate the generalization ability of HierDTA under both conventional and realistic drug discovery scenarios, experiments were conducted under four evaluation strategies.

\begin{itemize}
    \item \textbf{Warm setting.} Drug--target interaction pairs were randomly divided into training, validation, and test sets. Under this setting, both drugs and proteins appearing in the test set may also appear in the training set, while only interaction pairs are unseen.

    \item \textbf{Unseen drug setting.} The dataset was partitioned according to unique drug identifiers, ensuring that all drugs in the test set were completely absent from the training and validation sets. This setting evaluates the model's ability to generalize to novel compounds.

    \item \textbf{Unseen protein setting.} The dataset was partitioned according to unique protein identifiers so that proteins in the test set were never observed during training. This setting assesses the capability of predicting affinities for unseen targets.

    \item \textbf{All unseen setting.} Both drug and protein identifiers were simultaneously partitioned, ensuring that neither drugs nor proteins in the test set appeared during training or validation. This represents the most challenging cold-start scenario and closely resembles practical drug discovery tasks.
\end{itemize}

\subsection{Evaluation Metrics}

Following previous DTA studies, four widely adopted evaluation metrics were used to assess model performance, including Mean Squared Error (MSE), Concordance Index (CI), and $r_m^2$.

The MSE measures the average squared difference between predicted and true binding affinity values,

\begin{equation}
\mathrm{MSE}
=
\frac{1}{N}
\sum_{i=1}^{N}
(y_i-\hat{y}_i)^2,
\end{equation}

where $y_i$ and $\hat{y}_i$ denote the ground-truth and predicted affinity values, respectively.

The CI evaluates the ranking consistency between predicted and true affinities,

\begin{equation}
\mathrm{CI}
=
\frac{1}{Z}
\sum_{y_i>y_j}
h(\hat{y}_i-\hat{y}_j),
\end{equation}

where

\[
h(x)=
\begin{cases}
1,&x>0\\
0.5,&x=0\\
0,&x<0
\end{cases}
\]

and $Z$ denotes the normalization constant. Finally, the external predictive ability of the model is evaluated using the modified squared correlation coefficient,

\begin{equation}
r_m^2
=
r^2
\left(
1-\sqrt{|r^2-r_0^2|}
\right),
\end{equation}
wherer $r$ is the correlation coefficient with an intercept, and $r_0$ is the correlation coefficient without an intercept.

\subsection{Experimental Settings}
The proposed model was implemented using the PyTorch framework. All experiments were conducted on an NVIDIA A100 GPU with 40 GB memory. The model was optimized using the Adam optimizer with an initial learning rate of $1\times10^{-4}$. The maximum number of training epochs was set to 500, and early stopping was applied when the validation loss did not improve for 30 consecutive epochs. The batch size and other hyperparameters are summarized in Supporting Information S3. For a fair comparison, each experiment was repeated five times using different random seeds, and the mean and standard deviation were reported.

\section{Results and Discussion}
\textcolor{black}{To comprehensively evaluate the effectiveness of HierDTA, we compare it with representative DTA prediction models spanning sequence-based, graph-based, and recent multimodal frameworks, including DeepDTA~\cite{10.1093/bioinformatics/bty593}, GraphDTA~\cite{10.1093/bioinformatics/btaa921}, FusionDTA~\cite{yuan2022fusiondta}, MGraphDTA~\cite{D1SC05180F}, MSGNN-DTA~\cite{ijms24098326}, GRA-DTA~\cite{ijms25105126}, DMFF-DTA~\cite{he2025dual}, PMHGT-DTA~\cite{zhang2025multimodal}, and KANPM-DTA~\cite{rakib2026kanpm}}.

\subsection{Performance Comparison on Warm Settings}

\begin{figure*}[htbp]
    \centering
    \includegraphics[width=0.48\textwidth]{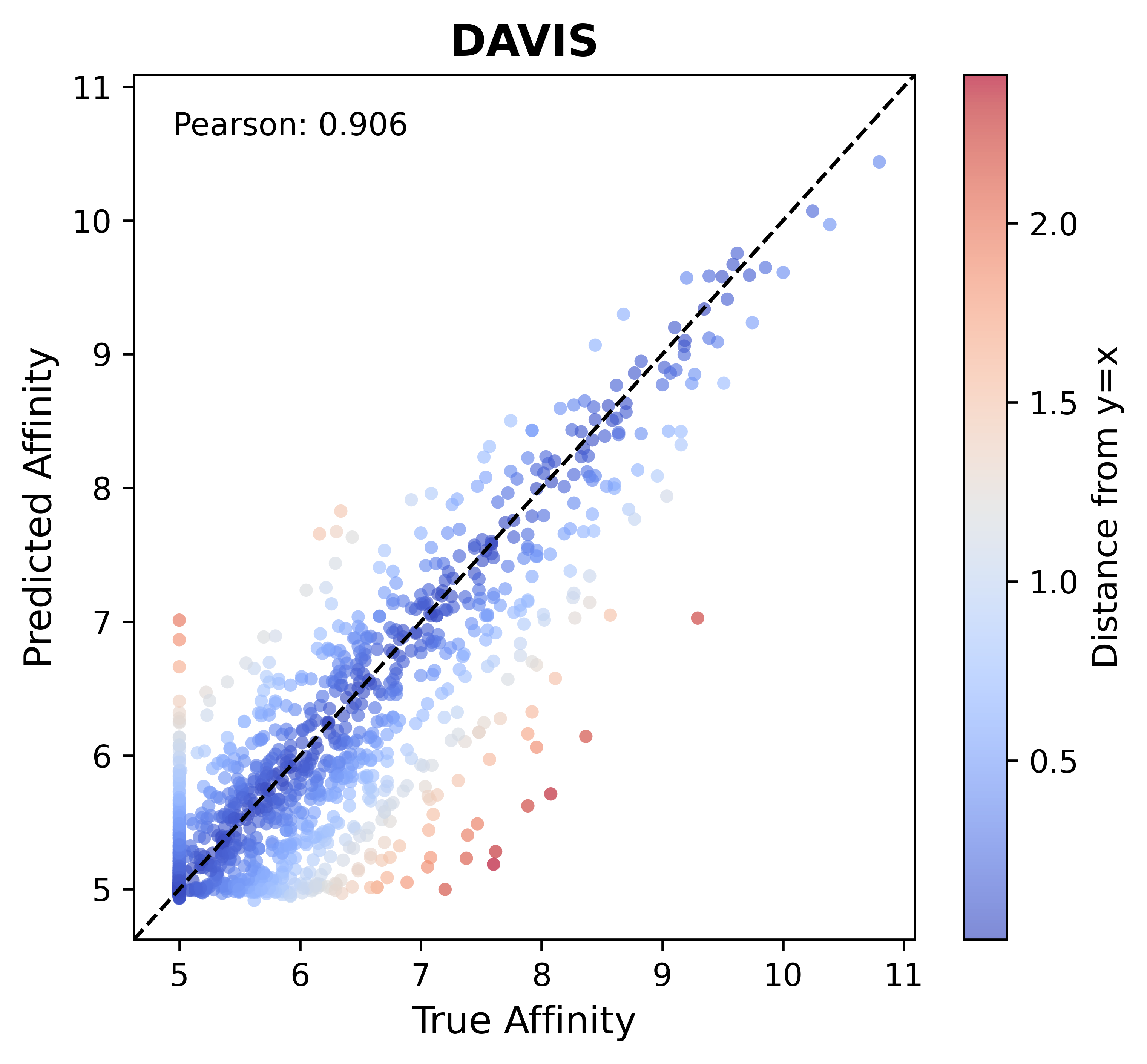}
    \hfill
    \includegraphics[width=0.48\textwidth]{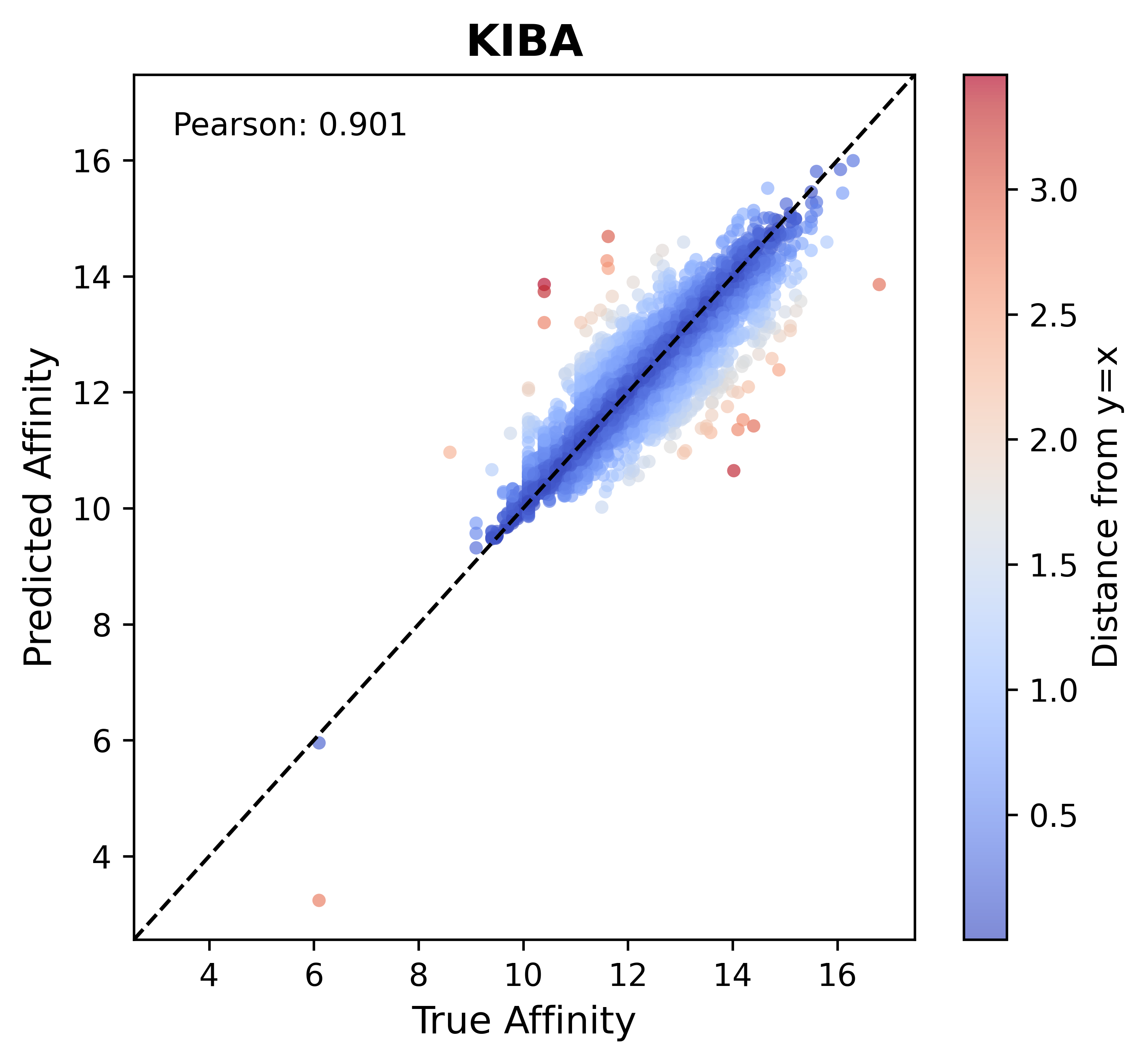}
    \caption{
    \textcolor{black}{Scatter plots of predicted vs true binding affinities on the Davis and KIBA test sets under the warm setting. Redder points indicate larger prediction errors, while blackr points indicate smaller errors.}
    }
    \label{fig:warm_scatter}
\end{figure*}

Table~\ref{tab:warm_performance} summarizes the performance of HierDTA and the baseline methods on the Davis and KIBA datasets under the warm setting. On Davis, HierDTA achieves the best result for all three evaluation metrics, with an MSE of \textcolor{black}{0.174, a CI of 0.905, and an $r_m^2$ of 0.757. Compared with KANPM-DTA, HierDTA reduces the MSE by approximately 14.7\% and improves the CI and $r_m^2$ by 0.8\% and 5.9\%, respectively}. These consistent improvements indicate that the proposed architecture effectively captures affinity-relevant information. On KIBA, HierDTA \textcolor{black}{achieves the highest \(r_m^2\) of 0.791, exceeding the value obtained by KANPM-DTA. Its MSE and CI remain close to the best values, achieved by KANPM-DTA and PMHGT-DTA, respectively. This may be partly attributed to the incorporation of pretrained node-level features in these competing methods, which can provide additional prior information for representation learning. The differences among the leading methods may also reflect the larger scale and broader interaction coverage of the KIBA dataset.} Overall, the results demonstrate that HierDTA remains competitive across datasets with distinct scales and affinity distributions. To further investigate the prediction behavior of HierDTA, Figure~\ref{fig:warm_scatter} visualizes the relationship between the predicted and true binding affinities on the Davis and KIBA test sets. Most sample points cluster near this diagonal, indicating high prediction accuracy. On the KIBA data set, scatter points concentrate more tightly along the diagonal, indicating better agreement between the model predictions and the ground-truth values. 

\begin{table*}[htbp]
\centering
\footnotesize
\setlength{\tabcolsep}{4.5pt}
\renewcommand{\arraystretch}{1.05}
\caption{Performance comparison under the warm setting on the Davis and KIBA datasets. The best result for each metric is shown in bold, and the second-best result is underlined.}
\label{tab:warm_performance}
\begin{tabular}{lcccccc}
\toprule
\multirow{2}{*}{Model}
& \multicolumn{3}{c}{Davis}
& \multicolumn{3}{c}{KIBA} \\
\cmidrule(lr){2-4} \cmidrule(lr){5-7}
& MSE$\downarrow$ & CI$\uparrow$ & $r_m^2\uparrow$
& MSE$\downarrow$ & CI$\uparrow$ & $r_m^2\uparrow$ \\
\midrule
DeepDTA
& $0.284\pm0.008$ & $0.869\pm0.002$ & $0.624\pm0.015$
& $0.201\pm0.003$ & $0.850\pm0.002$ & $0.671\pm0.017$ \\

GraphDTA
& $0.242\pm0.010$ & $0.880\pm0.001$ & $0.683\pm0.010$
& $0.192\pm0.001$ & $0.848\pm0.002$ & $0.718\pm0.004$ \\

FusionDTA
& $0.226\pm0.005$ & $0.891\pm0.001$ & $0.686\pm0.020$
& $0.155\pm0.002$ & $0.882\pm0.001$ & $0.750\pm0.015$ \\

MGraphDTA
& $0.228\pm0.004$ & $0.883\pm0.002$ & $0.679\pm0.022$
& $0.152\pm0.003$ & $0.884\pm0.002$ & $0.766\pm0.016$ \\

MSGNN-DTA
& $0.234\pm0.005$ & $0.890\pm0.002$ & $0.686\pm0.014$
& $0.149\pm0.003$ & $0.885\pm0.002$ & $0.763\pm0.009$ \\

DMFF-DTA
& $0.218\pm0.004$ & $0.894\pm0.002$ & $0.702\pm0.029$
& $0.144\pm0.002$ & $0.889\pm0.002$ & $0.773\pm0.016$ \\

PMHGT-DTA
& $0.205\pm0.010$ & $0.896\pm0.003$ & $\underline{0.726\pm0.019}$
& $0.142\pm0.003$ & $\mathbf{0.893\pm0.001}$ & $0.778\pm0.014$ \\

KANPM-DTA
& $\underline{0.204\pm0.005}$ & $\underline{0.898\pm0.001}$ & $0.715\pm0.015$
& $\mathbf{0.137\pm0.003}$ & $\underline{0.892\pm0.001}$ & $\underline{0.784\pm0.009}$ \\
HierDTA
& $\mathbf{0.174\pm0.016}$ & $\mathbf{0.905\pm0.004}$ & $\mathbf{0.757\pm0.026}$
& $\underline{0.140\pm0.008}$ & $0.888\pm0.004$ & $\mathbf{0.791\pm0.010}$ \\
\bottomrule
\end{tabular}

\vspace{1mm}
\begin{flushleft}

\end{flushleft}
\end{table*}

\subsection{Performance Comparison under Cold-start Settings}

\textcolor{black}{Standard warm splits on Davis and KIBA may lead to optimistic performance estimates because drugs or proteins in the test set can also appear in the training set through different interaction pairs. To mitigate this risk, we additionally evaluate HierDTA under more realistic scenarios, where the corresponding entities are explicitly held out from training.} Table~\ref{tab:cold_davis} and~\ref{tab:cold_kiba} compare the models under the three cold-start settings: unseen-drug, unseen-protein, and all-unseen. These settings provide a more stringent assessment of generalization than the warm setting.

On the Davis dataset, \textcolor{black}{the cold-start results reveal a clear setting-dependent generalization pattern of HierDTA. Under the unseen-drug setting, HierDTA achieves the highest \(r_m^2\), although its MSE and CI are less competitive than those of the strongest baselines. This discrepancy may be partly attributed to the limited number of unique drugs in Davis, which restricts the diversity of drug-side patterns available during training and makes extrapolation to unseen compounds particularly challenging. In contrast, HierDTA shows a clearer advantage when generalizing to unseen proteins, achieving the best CI and \(r_m^2\) while attaining the second-lowest MSE. This suggests that the architecture is particularly effective in capturing interaction patterns for previously unseen protein targets. The advantage becomes more pronounced in the all-unseen setting, where neither the drug nor the protein has been observed during training: HierDTA achieves the best CI and \(r_m^2\), with a particularly substantial gain in \(r_m^2\) over all competing methods. Taken together, these results indicate that HierDTA's main strength under cold-start conditions lies in preserving ranking consistency and predictive agreement for unseen entities, especially when protein-side generalization is required and under the most stringent all-unseen scenario.}

The results on the KIBA dataset further demonstrate the generalization capability of HierDTA under cold-start conditions. In the unseen-drug setting, HierDTA achieves the best performance across all three evaluation metrics, \textcolor{black}{reducing MSE by approximately 0.6\% and improving CI and \(r_m^2\) by 1.6\% and 10.7\%, respectively, compared with the second-best GRA-DTA. The advantage becomes more pronounced in the unseen-protein setting, where HierDTA reduces MSE by 21.7\% and improves CI and \(r_m^2\) by 0.7\% and 16.4\%, respectively, over DMFF-DTA.} These consistent improvements indicate that HierDTA generalizes effectively when either drugs or protein targets are unseen during training. 
Under the more challenging all-unseen setting, HierDTA improves CI and \(r_m^2\) by approximately \textcolor{black}{1.2\% and 31.8\% over the second-best results, respectively, although its MSE remains higher than that of GRA-DTA. One possible explanation is that GRA-DTA's globally aggregated representations and low-dimensional fusion may provide stronger implicit regularization, helping to suppress large absolute prediction errors under severe distribution shifts. In contrast, HierDTA's fine-grained hierarchical and interaction-level modeling may better preserve affinity ranking and overall predictive agreement, as reflected by its superior CI and \(r_m^2\).}

\textcolor{black}{
The cold-start generalization of HierDTA benefits from motif-to-atom augmentation for chemical representations, residue graph encoding for structure-aware protein representations, and global features that complement fine-grained interaction modeling. Its larger improvements on KIBA likely arise from the greater scale and diversity of the dataset, which facilitate the learning of more transferable drug and protein patterns.}

\begin{table*}[htbp]
\centering
\footnotesize
\renewcommand{\arraystretch}{1.08}
\caption{\textcolor{black}{Performance comparison under cold-start settings on the Davis dataset. The best results for each metric are shown in bold.}}
\label{tab:cold_davis}
\begin{tabular*}{\textwidth}{@{\extracolsep{\fill}}llccc@{}}
\toprule
\textbf{Setting} & \textbf{Model} & \textbf{MSE}$\downarrow$ & \textbf{CI}$\uparrow$ & $\boldsymbol{r_m^2}\uparrow$ \\
\midrule
\multirow{7}{*}{Unseen drug}
& GraphDTA    & $0.920\pm0.029$ & $0.678\pm0.036$ & $0.160\pm0.019$ \\
& FusionDTA   & $0.581\pm0.094$ & $0.737\pm0.012$ & $0.187\pm0.034$ \\
& MGraphDTA   & $0.563\pm0.065$ & $0.729\pm0.022$ & $0.192\pm0.021$ \\
& MSGNN-DTA   & $0.560\pm0.075$ & $0.731\pm0.036$ & $0.186\pm0.058$ \\
& GRA-DTA     & $0.578\pm0.005$ & $0.725\pm0.015$ & $0.121\pm0.024$ \\
& DMFF-DTA   & $\mathbf{0.548\pm0.063}$ & $\mathbf{0.742\pm0.037}$ & $0.228\pm0.073$ \\
& HierDTA & $0.637\pm0.095$ & $0.710\pm0.030$ & $\mathbf{0.232\pm0.081}$ \\
\addlinespace[2pt]
\midrule

\multirow{7}{*}{Unseen protein}
& GraphDTA   & $0.510\pm0.086$ & $0.729\pm0.012$ & $0.154\pm0.014$ \\
& FusionDTA   & $0.364\pm0.021$ & $0.826\pm0.011$ & $0.435\pm0.023$ \\
& MGraphDTA  & $0.359\pm0.023$ & $0.813\pm0.008$ & $0.425\pm0.028$ \\
& MSGNN-DTA  & $0.361\pm0.053$ & $0.816\pm0.018$ & $0.430\pm0.039$ \\
& GRA-DTA    & $0.376\pm0.006$ & $0.827\pm0.015$ & $0.453\pm0.024$ \\
& DMFF-DTA    & $\mathbf{0.330\pm0.047}$ & $0.840\pm0.014$ & $0.501\pm0.044$ \\
& HierDTA & $0.347\pm0.049$ & $\mathbf{0.854\pm0.016}$ & $\mathbf{0.551\pm0.069}$ \\
\addlinespace[2pt]
\midrule

\multirow{7}{*}{All unseen}
& GraphDTA   & $0.968\pm0.096$ & $0.579\pm0.017$ & $0.026\pm0.016$ \\
& FusionDTA  & $0.876\pm0.091$ & $0.645\pm0.043$ & $0.072\pm0.048$ \\
& MGraphDTA   & $0.874\pm0.090$ & $0.636\pm0.021$ & $0.071\pm0.041$ \\
& MSGNN-DTA   & $0.791\pm0.078$ & $0.650\pm0.053$ & $0.076\pm0.064$ \\
& GRA-DTA     & $\mathbf{0.560\pm0.007}$ & $0.690\pm0.012$ & $0.101\pm0.024$ \\
& DMFF-DTA    & $0.759\pm0.071$ & $0.655\pm0.050$ & $0.095\pm0.076$ \\
& HierDTA & $0.606\pm0.114$ & $\mathbf{0.704\pm0.057}$ & $\mathbf{0.237\pm0.075}$ \\
\bottomrule
\end{tabular*}

\vspace{1mm}

\end{table*}

\begin{table*}[htbp]
\centering
\footnotesize
\renewcommand{\arraystretch}{1.08}
\caption{\textcolor{black}{Performance comparison under cold-start settings on the KIBA dataset. The best results for each metric are shown in bold.}}
\label{tab:cold_kiba}
\begin{tabular*}{\textwidth}{@{\extracolsep{\fill}}llccc@{}}
\toprule
\textbf{Setting} & \textbf{Model} & \textbf{MSE}$\downarrow$ & \textbf{CI}$\uparrow$ & $\boldsymbol{r_m^2}\uparrow$ \\
\midrule
\multirow{7}{*}{Unseen drug}
& GraphDTA   & $0.471\pm0.047$ & $0.713\pm0.002$ & $0.342\pm0.007$ \\
& FusionDTA   & $0.429\pm0.031$ & $0.748\pm0.005$ & $0.364\pm0.012$ \\
& MGraphDTA  & $0.425\pm0.047$ & $0.746\pm0.002$ & $0.366\pm0.016$ \\
& MSGNN-DTA   & $0.426\pm0.043$ & $0.747\pm0.003$ & $0.358\pm0.022$ \\
& GRA-DTA     & $0.337\pm0.007$ & $0.765\pm0.012$ & $0.466\pm0.024$ \\
& DMFF-DTA    & $0.408\pm0.036$ & $0.753\pm0.009$ & $0.397\pm0.020$ \\
& HierDTA & $\mathbf{0.335\pm0.014}$ & $\mathbf{0.777\pm0.005}$ & $\mathbf{0.516\pm0.018}$ \\
\addlinespace[2pt]
\midrule

\multirow{7}{*}{Unseen protein}
& GraphDTA   & $0.469\pm0.089$ & $0.610\pm0.035$ & $0.368\pm0.057$ \\
& FusionDTA  & $0.439\pm0.062$ & $0.685\pm0.032$ & $0.390\pm0.067$ \\
& MGraphDTA  & $0.435\pm0.055$ & $0.674\pm0.028$ & $0.382\pm0.047$ \\
& MSGNN-DTA   & $0.438\pm0.061$ & $0.683\pm0.025$ & $0.399\pm0.054$ \\
& GRA-DTA     & $0.435\pm0.007$ & $0.692\pm0.012$ & $0.355\pm0.024$ \\
& DMFF-DTA   & $0.410\pm0.063$ & $0.748\pm0.053$ & $0.446\pm0.064$ \\
& HierDTA & $\mathbf{0.321\pm0.082}$ & $\mathbf{0.753\pm0.024}$ & $\mathbf{0.519\pm0.086}$ \\
\addlinespace[2pt]
\midrule

\multirow{7}{*}{All unseen}
& GraphDTA    & $0.676\pm0.113$ & $0.601\pm0.030$ & $0.149\pm0.067$ \\
& FusionDTA  & $0.587\pm0.086$ & $0.641\pm0.023$ & $0.193\pm0.053$ \\
& MGraphDTA  & $0.590\pm0.094$ & $0.626\pm0.028$ & $0.182\pm0.012$ \\
& MSGNN-DTA   & $0.581\pm0.079$ & $0.648\pm0.038$ & $0.180\pm0.021$ \\
& GRA-DTA     & $\mathbf{0.441\pm0.007}$ & $0.657\pm0.012$ & $0.228\pm0.024$ \\
& DMFF-DTA    & $0.567\pm0.082$ & $0.667\pm0.035$ & $0.236\pm0.037$ \\
& HierDTA & $0.511\pm0.042$ & $\mathbf{0.675\pm0.025}$ & $\mathbf{0.311\pm0.055}$ \\
\bottomrule
\end{tabular*}

\vspace{1mm}

\end{table*}

\subsection{Ablation Study}
To examine the effects of the principal architectural choices, we conducted several ablation experiments on Davis. \textcolor{black}{In the drug encoder comparison, the same gnn type was used for both the atom and motif branches, while the remaining model configuration was held fixed. For the component ablation, two variants were evaluated. ``No Augmentation'' removes the motif-to-atom augmentation, leaving only the atom-level drug representation. ``No Interaction'' removes the cross-attention module, thereby eliminating explicit interaction between the drug and protein representations. These variants allow the contributions of hierarchical molecular representation and drug-protein interaction modeling to be assessed separately. The results are summarized in Table~\ref{tab:ablation}.}

\begin{table*}[htbp]
\centering
\footnotesize
\setlength{\tabcolsep}{5pt}
\renewcommand{\arraystretch}{1.08}
\caption{\textcolor{black}{Ablation results on the Davis dataset. The best result within each panel is shown in bold.}}
\label{tab:ablation}
\begin{tabular}{lccc}
\toprule
\textbf{Configuration} & \textbf{MSE}$\downarrow$ & \textbf{CI}$\uparrow$ & $\boldsymbol{r_m^2}\uparrow$ \\
\midrule
\multicolumn{4}{l}{\textbf{(A) Drug graph encoder}} \\
GAT & $\mathbf{0.174\pm0.016}$ & $\mathbf{0.905\pm0.004}$ & $\mathbf{0.757\pm0.026}$ \\
GCN   & $0.183\pm0.029$ & $0.899\pm0.005$ & $0.744\pm0.040$ \\
GIN   & $0.184\pm0.022$ & $0.903\pm0.005$ & $0.756\pm0.027$ \\
\midrule
\multicolumn{4}{l}{\textbf{(B) Component}} \\
HierDTA &  $\mathbf{0.174\pm0.016}$ & $\mathbf{0.905\pm0.004}$ & $\mathbf{0.757\pm0.026}$ \\
No Augmentation & $0.187\pm0.026$ & $0.900\pm0.004$ & $0.741\pm0.046$ \\
No Interaction  & $0.196\pm0.025$ & $0.903\pm0.006$ & $0.743\pm0.030$ \\
\bottomrule
\end{tabular}
\end{table*}

\textcolor{black}{As shown in panel (A) of Table~\ref{tab:ablation}, GAT achieved the lowest MSE and the highest CI and $r_m^2$ among the three evaluated encoders. Relative to GCN and GIN, GAT reduced the MSE by 4.9\% and 5.4\%, respectively. Overall, GAT provided the most favorable balance across the evaluated metrics and was therefore adopted as the drug graph encoder in the final configuration. The component ablation in Panel (B) of Table~\ref{tab:ablation} further demonstrates the contribution of hierarchical molecular representation and interaction modeling. When the motif-to-atom augmentation is removed and only atom-level representations are retained, the MSE increases from 0.174 to 0.187, while CI and \(r_m^2\) decrease from 0.905 and 0.757 to 0.900 and 0.741, respectively. These consistent degradations indicate that incorporating motif-level information provides complementary structural context beyond atom-level features and improves affinity prediction. Removing the interaction module leads to a further increase in MSE to 0.196 and a decrease in \(r_m^2\) to 0.743, highlighting the importance of explicitly modeling drug–target interactions. Together, these results support the complementary roles of hierarchical molecular representation and interaction modeling in HierDTA.}

\begin{figure*}[htbp]
    \centering
    \includegraphics[width=\textwidth]{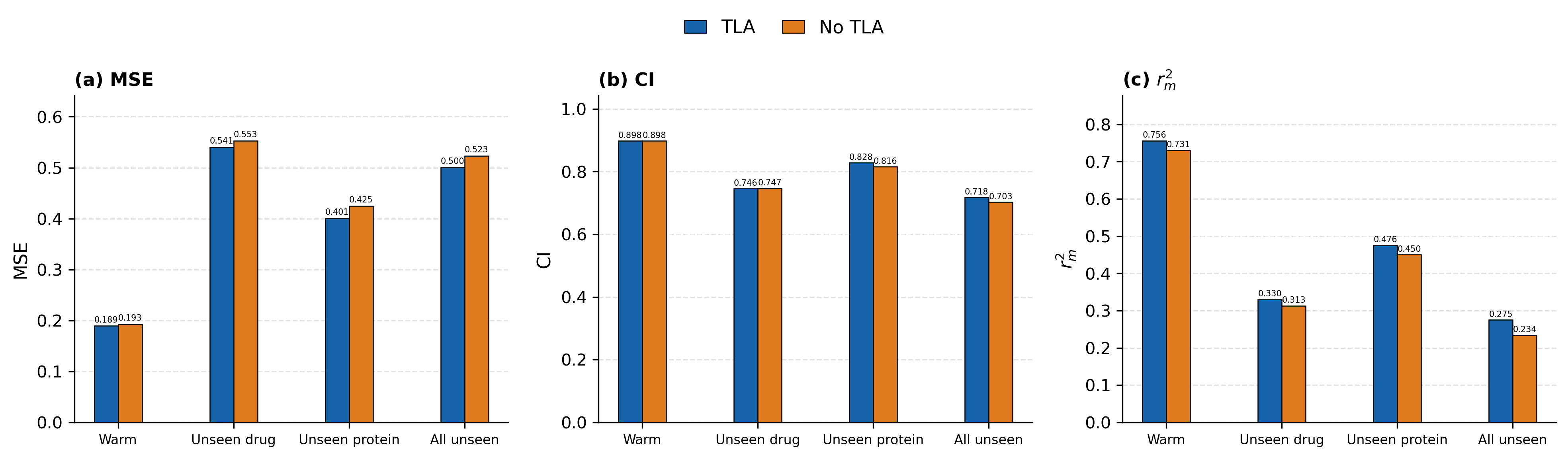}
    \caption{\textcolor{black}{Comparison of the TLA and No TLA operation on the Davis dataset across the four evaluation settings: (a) MSE, (b) CI, and (c) $r_m^2$. Lower MSE and higher CI and $r_m^2$ values are better.}}
    \label{fig:tla_vs_notla}
\end{figure*}

\textcolor{black}{We further compared the attention-based TLA operation with a simple variant in which the motif context is injected into the atom messages directly as a residual term (No TLA). Figure~\ref{fig:tla_vs_notla} reports the comparison across the four evaluation settings generated from a single random seed. Under the warm setting, TLA attains a lower MSE and a substantially higher $r_m^2$. The most pronounced improvements appear under the unseen-protein and all-unseen settings, where TLA improves all three metrics. These results indicate that the attention-based method in TLA supports effective information injection to atoms when the model must generalize beyond the training distribution.}

\subsection{Interpretability Analysis}
\label{sec:interpret}
\begin{figure*}[htbp]
    \centering
    \includegraphics[width=\textwidth]{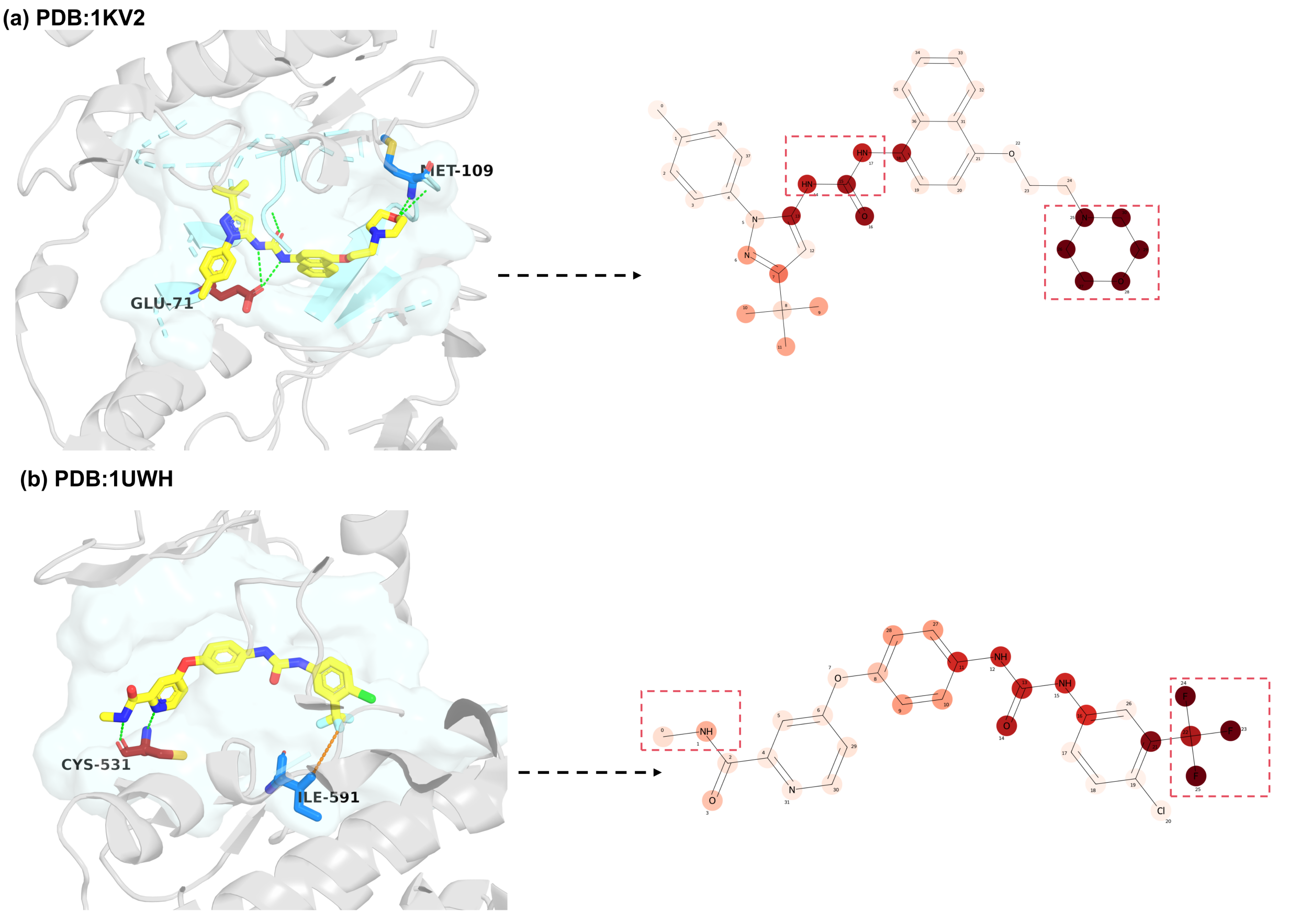}
    \caption{\textcolor{black}{Interpretability analysis of drug-protein complexes. (a) 1KV2 complex: The left panel shows the experimentally resolved 3D binding conformation, with key residues assigned high attention weights by HierDTA labeled. The right panel shows the atom-level attention heatmap over the ligand (darker colors indicate higher weights). (b) 1UWH complex: All subpanels have the same meanings as those in (a).}}
    \label{fig:attention_map}
\end{figure*}

\textcolor{black}{
HierDTA incorporates a cross-attention module that enables atom--residue interactions to be interpreted through attention weights. To assess whether the learned representations capture chemically and biologically relevant regions, we analyzed two experimentally resolved drug--protein complexes (PDB: 1KV2 and 1UWH), and compared the attention patterns with their binding conformations. As shown in Figure~\ref{fig:attention_map}, the prominent features on the protein and ligand sides exhibit noticeable agreement. For 1KV2 (Figure~\ref{fig:attention_map}(a)), residues GLU-71 and MET-109 receive relatively high attention, while highly weighted ligand atoms are concentrated in nearby molecular substructures (shown in the red dashed boxes). A similar pattern is observed for 1UWH (Figure~\ref{fig:attention_map}(b)), where CYS-531 and ILE-591 are associated with ligand regions showing elevated atom attention. However, some model predictions do not fully agree with the experimentally resolved interactions. This is reasonable because the attention weights are learned from the DTA prediction objective without explicit supervision from atom--residue contact annotations. Thus, the attention maps should be used only as a qualitative reference for interpreting potentially interaction-relevant regions.}

\begin{figure*}[htbp]
    \centering
    \includegraphics[width=\textwidth]{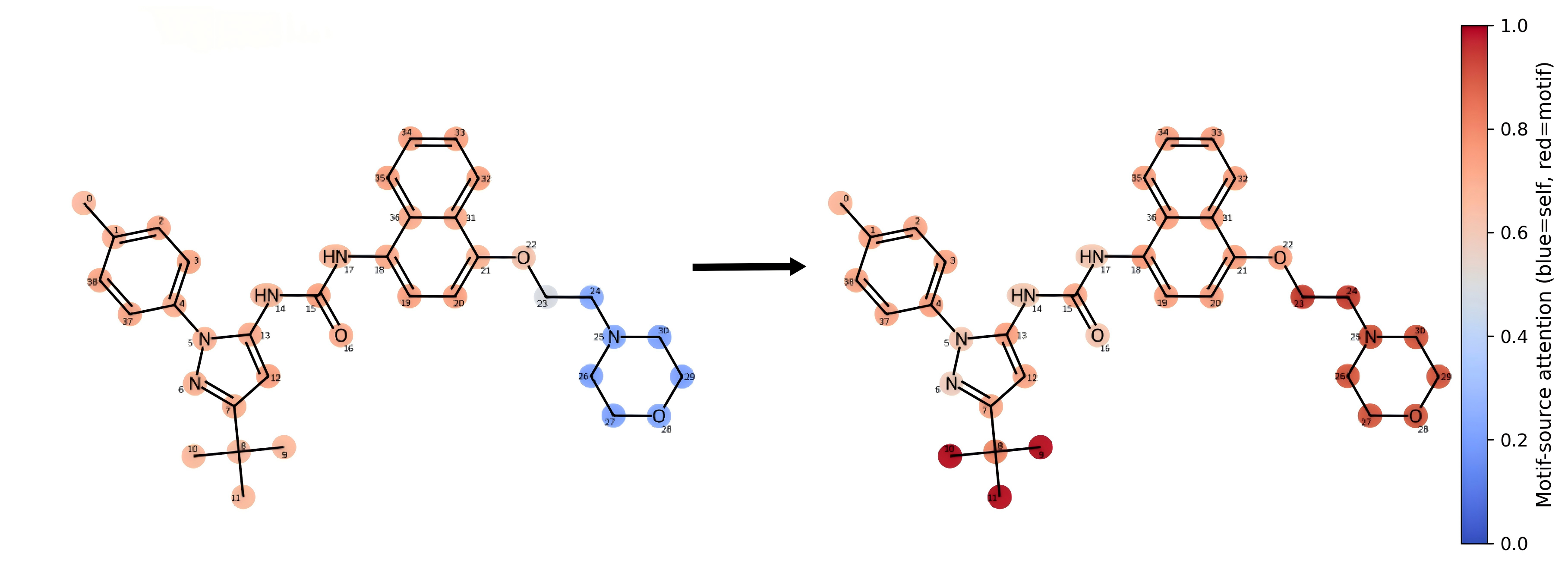}
    \caption{\textcolor{black}{Motif attention weights from the first and second TLA operations for 1KV2 ligand. Values closer to one indicate greater reliance on the motif-derived context, whereas values closer to zero indicate greater reliance on the atom structural message.}}
    \label{fig:motif_weight}
\end{figure*}

\textcolor{black}{
Figure~\ref{fig:motif_weight} further illustrates how motif information is incorporated at different depths of the drug encoder, using ligand B96 in the 1KV2 complex as an example. The visualized weights indicate the contribution of motif-level information to each atom during augmentation. In the first TLA operation, motif weights remain relatively low across the ligand, indicating a limited contribution of motif features at this stage. In the second operation, the motif contribution becomes more localized, with particularly large weights assigned to the terminal heterocyclic group and the branched substituent. The distribution suggests that the augmentation operations are not redundant. Instead, as the augmentation proceeds, motif information becomes increasingly selective and localized to specific atoms within important functional groups.}

\section{Case Study}
To further evaluate the practical utility of HierDTA in drug screening scenarios, we conducted two case studies. Case study 1 evaluates the predicted affinity ranking and docking scores of DrugBank compounds for EGFR, whereas case study 2 evaluates early active-compound enrichment for EGFR and VEGFR2 relative to MSGNN-DTA~\cite{ijms24098326}.

\textit{Case study 1: EGFR--DrugBank screening.} EGFR (UniProt ID: P00533), a clinically important kinase target~\cite{sigismund2018emerging}, was screened against 3136 DrugBank~\cite{knox2024drugbank} small molecules excluded from the KIBA dataset. Compounds were ranked in descending order of their HierDTA-predicted KIBA scores. Duplicate representations of the same parent compound were consolidated before reporting. The ten highest-ranked unique compounds are summarized in Table~\ref{tab:case_study}. \textcolor{black}{Five established EGFR inhibitors (dacomitinib, canertinib, afatinib, poziotinib, and gefitinib) were recovered among the top candidates. Molecular docking was carried out with CB-Dock2~\cite{liu2022cb} to assess the binding plausibility of these candidates. The Vina scores of the ten compounds ranged from $-10.2$ to $-7.9$ kcal/mol. Dacomitinib, the highest-ranked compound, also showed the most favorable docking score ($-10.2$~kcal/mol)}, although the HierDTA ranking and docking ranking were not fully consistent for the remaining compounds. This is reasonable because the two methods evaluate binding from different perspectives: HierDTA predicts affinity from learned drug and protein representations, whereas CB-Dock2 scores a generated 3D binding pose using an empirical scoring function.
\textcolor{black}{We further examined two top-ranked candidates, genistein and emodin, whose EGFR binding affinities were not available in our dataset, together with dacomitinib as a reference. Both candidates can be accommodated in the ATP-binding pocket. Their relatively rigid and planar scaffolds are compatible with hydrophobic packing, while their carbonyl and hydroxyl groups provide possible hydrogen-bonding sites with surrounding residues. The docking poses of genistein, emodin, and dacomitinib in the EGFR ATP-binding pocket are shown in the Supporting Information S5. These features provide a structural rationale for their high predicted rankings, but the docking results should be viewed as supporting computational evidence rather than experimental validation. Potential false-positive predictions may arise from several sources, including biases in the training data, approximations in the protein--ligand conformations used for modeling, the omission of solvent effects, binding dynamics, and the cellular environment. The candidates identified here should be considered priorities for further experimental testing rather than confirmed EGFR binders or inhibitors.}

\begin{table*}[htbp]
\centering
\footnotesize
\setlength{\tabcolsep}{12pt}
\renewcommand{\arraystretch}{1.10}
\caption{\textcolor{black}{Top-10 unique DrugBank compounds ranked by HierDTA for EGFR (PDB ID: 3POZ). Boldface denotes established EGFR inhibitors.}}
\label{tab:case_study}

\begin{tabular}{c l c c}
\toprule
Rank & Compound & Predicted KIBA & Vina Score (kcal/mol)\\
\midrule
1  & \textbf{Dacomitinib} & 13.912 & $-10.2$\\
2  & \textbf{Canertinib} & 13.801 & $-9.2$\\
3  & JAK3-inhibitor-I & 13.792 & $-7.9$\\
4  & \textbf{Afatinib} & 13.786 & $-8.9$\\
5  & \textbf{Poziotinib} & 13.755 & $-9.5$\\
6  & \textbf{Gefitinib} & 13.696 & $-8.6$\\
7  & Prunetin & 13.401 & $-8.2$\\
8  & Genistein & 13.342 & $-8.3$\\
9  & Emodin & 13.309 & $-8.8$\\
10 & Staurosporine & 13.269 & $-9.6$\\
\bottomrule
\end{tabular}
\end{table*}

\begin{figure*}[htbp]
    \centering
    \includegraphics[width=\textwidth]{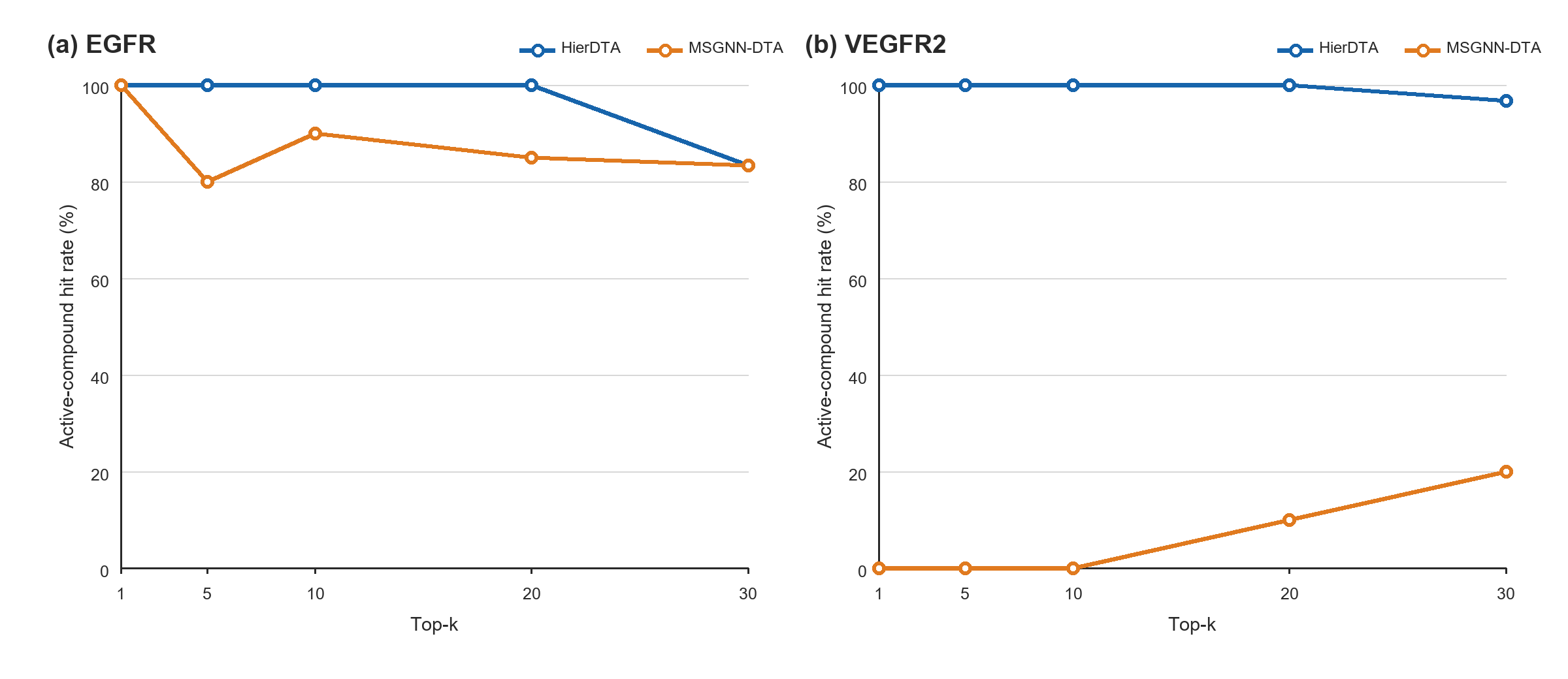}
    \caption{\textcolor{black}{Top-$k$ active compound hit rates of HierDTA and MSGNN-DTA on the external ChEMBL evaluation sets for (a) EGFR and (b) VEGFR2.}}
    \label{fig:egfr_vegfr2_topk}
\end{figure*}

\textit{Case study 2: early enrichment of EGFR and VEGFR2 active compounds.} We next assessed whether experimentally active compounds were enriched at the top of EGFR and VEGFR2 rankings. Quantitative IC$_{50}$ records were obtained from ChEMBL~\cite{zdrazil2024chembl}, and replicate measurements for each compound--target pair were summarized by the median. Compounds with median IC$_{50}\leq1$ nM were designated active, whereas those with median IC$_{50}\geq10^5$ nM were designated inactive; intermediate cases were excluded. For each target, we constructed a balanced evaluation set containing 50 active and 50 inactive compounds collected from ChEMBL, which are external to the KIBA dataset. HierDTA and MSGNN-DTA~\cite{ijms24098326} independently ranked the same target-specific set according to their predicted KIBA scores. The top-$k$ hit rate was defined as the proportion of active compounds among the first $k$ ranked candidates. As shown in Figure~\ref{fig:egfr_vegfr2_topk}, HierDTA achieved stronger early enrichment than MSGNN-DTA on both the EGFR and VEGFR2 sets, indicating that active compounds were more frequently prioritized near the top of the ranking. \textcolor{black}{This experiment is intended to assess retrospective ranking performance under the evaluation set. Although a higher top-$k$ hit rate reflects better early enrichment, it does not eliminate false-positive predictions in practical screening. The results should therefore be interpreted as evidence of relative ranking ability rather than direct validation of compound activity.}

\section{Conclusions}
In this study, we proposed HierDTA, a hierarchical structure-aware framework for drug--target affinity prediction that integrates multi-level drug representation, protein residue graph learning, and drug--target interaction modeling. \textcolor{black}{The core idea is to propagate functional-motif
semantics to atom-level representations, encode residue interaction graphs
with edge-aware message passing, and integrate both through cross-attention.} Experiments on the Davis and KIBA datasets demonstrated that HierDTA consistently achieves competitive performance against state-of-the-art methods under both warm and cold-start settings. \textcolor{black}{Under the warm setting, HierDTA achieved the best overall performance on Davis (MSE: 0.174, CI: 0.905, \(r_m^2\): 0.757) and the highest \(r_m^2\) of 0.791 on KIBA. Moreover, HierDTA maintained strong performance under the more challenging cold-start settings, demonstrating its generalization capability to more realistic scenarios. Ablation studies supported the contributions of the augmentation and interaction components.} Case studies on EGFR and VEGFR2 further demonstrated that HierDTA can provide effective early enrichment of candidate bioactive compounds, suggesting its potential utility for virtual screening. 

\textcolor{black}{However, several limitations remain. First, the functional motifs used in this work are identified using predefined rules. Although such motifs provide chemically interpretable structural priors, they may not capture biological functional unit and previously unseen chemical scaffolds. Second, although the cold-start settings provide more challenging evaluations than warm split, similarity-based overlap may still remain between training and test sets. Future work will explore data-driven motif discovery and consider scaffold- or similarity-based entity splits under evaluation settings.}

\section{Acknowledgments}
This work was supported by the Noncommunicable Chronic Diseases-National Science and Technology Major Project (2024ZD0533500).

\section{Data and Software Availability}
The data and source codes are available on \href{https://github.com/xuhuangchao/HierDTA.git}{github}.

\section{Supporting Information}
Rule-based motif decomposition and representative examples, details of protein structure preprocessing and residue-graph construction, hyperparameter settings, comparison of protein encoders, and molecular docking poses of candidate compounds with EGFR.

\section{Author Contributions}
Huangchao Xu conceived and conducted the experiments, Huangchao Xu and Baohua Zhang wrote and reviewed the manuscript, Qian Liu, Zhong Jin and Yi Du supervised the study and provided technical support. 

\printbibliography

@article{10.1093/bioinformatics/bty593,
    author = {Öztürk, Hakime and Özgür, Arzucan and Ozkirimli, Elif},
    title = {DeepDTA: deep drug–target binding affinity prediction},
    journal = {Bioinformatics},
    volume = {34},
    number = {17},
    pages = {i821-i829},
    year = {2018},
    month = {09},
    issn = {1367-4803},
    doi = {10.1093/bioinformatics/bty593},
    url = {https://doi.org/10.1093/bioinformatics/bty593},
    eprint = {https://academic.oup.com/bioinformatics/article-pdf/34/17/i821/50582374/bioinformatics_34_17_i821.pdf},
}

@article{10.1093/bioinformatics/btaa921,
    author = {Nguyen, Thin and Le, Hang and Quinn, Thomas P and Nguyen, Tri and Le, Thuc Duy and Venkatesh, Svetha},
    title = {GraphDTA: predicting drug–target binding affinity with graph neural networks},
    journal = {Bioinformatics},
    volume = {37},
    number = {8},
    pages = {1140-1147},
    year = {2020},
    month = {10},
    issn = {1367-4803},
    doi = {10.1093/bioinformatics/btaa921},
    url = {https://doi.org/10.1093/bioinformatics/btaa921},
    eprint = {https://academic.oup.com/bioinformatics/article-pdf/37/8/1140/50340643/btaa921.pdf},
}

@Article{D0RA02297G,
author ="Jiang, Mingjian and Li, Zhen and Zhang, Shugang and Wang, Shuang and Wang, Xiaofeng and Yuan, Qing and Wei, Zhiqiang",
title  ="Drug–target affinity prediction using graph neural network and contact maps",
journal  ="RSC Adv.",
year  ="2020",
volume  ="10",
issue  ="35",
pages  ="20701-20712",
publisher  ="The Royal Society of Chemistry",
doi  ="10.1039/D0RA02297G",
url  ="http://dx.doi.org/10.1039/D0RA02297G"}

@Article{D1SC05180F,
author ="Yang, Ziduo and Zhong, Weihe and Zhao, Lu and Yu-Chian Chen, Calvin",
title  ="MGraphDTA: deep multiscale graph neural network for explainable drug–target binding affinity prediction",
journal  ="Chem. Sci.",
year  ="2022",
volume  ="13",
issue  ="3",
pages  ="816-833",
publisher  ="The Royal Society of Chemistry",
doi  ="10.1039/D1SC05180F",
url  ="http://dx.doi.org/10.1039/D1SC05180F"}

@article{10.1109/TCBB.2022.3170365,
author = {Zhao, Qichang and Duan, Guihua and Yang, Mengyun and Cheng, Zhongjian and Li, Yaohang and Wang, Jianxin},
title = {AttentionDTA: Drug–Target Binding Affinity Prediction by Sequence-Based Deep Learning With Attention Mechanism},
year = {2022},
issue_date = {March-April 2023},
publisher = {IEEE Computer Society Press},
address = {Washington, DC, USA},
volume = {20},
number = {2},
issn = {1545-5963},
url = {https://doi.org/10.1109/TCBB.2022.3170365},
doi = {10.1109/TCBB.2022.3170365},
journal = {IEEE/ACM Trans. Comput. Biol. Bioinformatics},
month = apr,
pages = {852–863},
numpages = {12}
}

@article{WU2024623,
title = {AttentionMGT-DTA: A multi-modal drug-target affinity prediction using graph transformer and attention mechanism},
journal = {Neural Networks},
volume = {169},
pages = {623-636},
year = {2024},
issn = {0893-6080},
doi = {https://doi.org/10.1016/j.neunet.2023.11.018},
url = {https://www.sciencedirect.com/science/article/pii/S089360802300641X},
author = {Hongjie Wu and Junkai Liu and Tengsheng Jiang and Quan Zou and Shujie Qi and Zhiming Cui and Prayag Tiwari and Yijie Ding}
}

@Article{D3RA00281K,
author ="Voitsitskyi, Taras and Stratiichuk, Roman and Koleiev, Ihor and Popryho, Leonid and Ostrovsky, Zakhar and Henitsoi, Pavlo and Khropachov, Ivan and Vozniak, Volodymyr and Zhytar, Roman and Nechepurenko, Diana and Yesylevskyy, Semen and Nafiiev, Alan and Starosyla, Serhii",
title  ="3DProtDTA: a deep learning model for drug-target affinity prediction based on residue-level protein graphs",
journal  ="RSC Adv.",
year  ="2023",
volume  ="13",
issue  ="15",
pages  ="10261-10272",
publisher  ="The Royal Society of Chemistry",
doi  ="10.1039/D3RA00281K",
url  ="http://dx.doi.org/10.1039/D3RA00281K"}

@article{zhao2024pocketdta,
  title={PocketDTA: an advanced multimodal architecture for enhanced prediction of drug- target affinity from 3D structural data of target binding pockets},
  author={Zhao, Long and Wang, Hongmei and Shi, Shaoping},
  journal={Bioinformatics},
  volume={40},
  number={10},
  pages={btae594},
  year={2024},
  publisher={Oxford University Press}
}

@article{yuan2022fusiondta,
  title={FusionDTA: attention-based feature polymerizer and knowledge distillation for drug-target binding affinity prediction},
  author={Yuan, Weining and Chen, Guanxing and Chen, Calvin Yu-Chian},
  journal={Briefings in Bioinformatics},
  volume={23},
  number={1},
  pages={bbab506},
  year={2022},
  publisher={Oxford University Press}
}

@article{rives2021biological,
  title={Biological structure and function emerge from scaling unsupervised learning to 250 million protein sequences},
  author={Rives, Alexander and Meier, Joshua and Sercu, Tom and Goyal, Siddharth and Lin, Zeming and Liu, Jason and Guo, Demi and Ott, Myle and Zitnick, C Lawrence and Ma, Jerry and others},
  journal={Proceedings of the national academy of sciences},
  volume={118},
  number={15},
  pages={e2016239118},
  year={2021},
  publisher={National Academy of Sciences}
}

@article{lin2023evolutionary,
  title={Evolutionary-scale prediction of atomic-level protein structure with a language model},
  author={Lin, Zeming and Akin, Halil and Rao, Roshan and Hie, Brian and Zhu, Zhongkai and Lu, Wenting and Smetanin, Nikita and Verkuil, Robert and Kabeli, Ori and Shmueli, Yaniv and others},
  journal={Science},
  volume={379},
  number={6637},
  pages={1123--1130},
  year={2023},
  publisher={American Association for the Advancement of Science}
}

@article{he2023nhgnn,
  title={NHGNN-DTA: a node-adaptive hybrid graph neural network for interpretable drug--target binding affinity prediction},
  author={He, Haohuai and Chen, Guanxing and Chen, Calvin Yu-Chian},
  journal={Bioinformatics},
  volume={39},
  number={6},
  pages={btad355},
  year={2023},
  publisher={Oxford University Press}
}

@article{davis2011comprehensive,
  title={Comprehensive analysis of kinase inhibitor selectivity},
  author={Davis, Mindy I and Hunt, Jeremy P and Herrgard, Sanna and Ciceri, Pietro and Wodicka, Lisa M and Pallares, Gabriel and Hocker, Michael and Treiber, Daniel K and Zarrinkar, Patrick P},
  journal={Nature biotechnology},
  volume={29},
  number={11},
  pages={1046--1051},
  year={2011},
  publisher={Nature Publishing Group US New York}
}

@article{tang2014making,
  title={Making sense of large-scale kinase inhibitor bioactivity data sets: a comparative and integrative analysis},
  author={Tang, Jing and Szwajda, Agnieszka and Shakyawar, Sushil and Xu, Tao and Hintsanen, Petteri and Wennerberg, Krister and Aittokallio, Tero},
  journal={Journal of chemical information and modeling},
  volume={54},
  number={3},
  pages={735--743},
  year={2014},
  publisher={ACS Publications}
}

@article{brody2021attentive,
  title={How attentive are graph attention networks?},
  author={Brody, Shaked and Alon, Uri and Yahav, Eran},
  journal={arXiv preprint arXiv:2105.14491},
  year={2021}
}

@article{rogers2010extended,
  title={Extended-connectivity fingerprints},
  author={Rogers, David and Hahn, Mathew},
  journal={Journal of chemical information and modeling},
  volume={50},
  number={5},
  pages={742--754},
  year={2010},
  publisher={ACS Publications}
}

@Article{ijms24098326,
AUTHOR = {Wang, Shudong and Song, Xuanmo and Zhang, Yuanyuan and Zhang, Kuijie and Liu, Yingye and Ren, Chuanru and Pang, Shanchen},
TITLE = {MSGNN-DTA: Multi-Scale Topological Feature Fusion Based on Graph Neural Networks for Drug–Target Binding Affinity Prediction},
JOURNAL = {International Journal of Molecular Sciences},
VOLUME = {24},
YEAR = {2023},
NUMBER = {9},
ARTICLE-NUMBER = {8326},
URL = {https://www.mdpi.com/1422-0067/24/9/8326},
PubMedID = {37176031},
ISSN = {1422-0067},
DOI = {10.3390/ijms24098326}
}

@Article{ijms25105126,
AUTHOR = {Tang, Xin and Lei, Xiujuan and Zhang, Yuchen},
TITLE = {Prediction of Drug-Target Affinity Using Attention Neural Network},
JOURNAL = {International Journal of Molecular Sciences},
VOLUME = {25},
YEAR = {2024},
NUMBER = {10},
ARTICLE-NUMBER = {5126},
URL = {https://www.mdpi.com/1422-0067/25/10/5126},
PubMedID = {38791165},
ISSN = {1422-0067},
DOI = {10.3390/ijms25105126}
}

@article{xu2025mmsg,
  title={MMSG-DTA: A multimodal, multiscale model based on sequence and graph modalities for drug-target affinity prediction},
  author={Xu, Jiahao and Ci, Lei and Zhu, Bo and Zhang, Guanhua and Jiang, Linhua and Ye-Lehmann, Shixin and Long, Wei},
  journal={Journal of Chemical Information and Modeling},
  volume={65},
  number={2},
  pages={981--996},
  year={2025},
  publisher={ACS Publications}
}

@article{he2025dual,
  title={Dual modality feature fused neural network integrating binding site information for drug target affinity prediction},
  author={He, Haohuai and Chen, Guanxing and Tang, Zhenchao and Chen, Calvin Yu-Chian},
  journal={NPJ Digital Medicine},
  volume={8},
  number={1},
  pages={67},
  year={2025},
  publisher={Nature Publishing Group UK London}
}

@article{rakib2026kanpm,
  title={KANPM-DTA: improving drug--target affinity prediction with Kolmogorov--Arnold networks and pretrained models},
  author={Rakib, MD Youshuf Khan and Alamin, Muhammad Habibulla and Li, Jiamu and Mamun, Sheikh Sohan and Gobena, Kaleb Amsalu and Ren, Shengbing},
  journal={Briefings in Bioinformatics},
  volume={27},
  number={2},
  pages={bbag112},
  year={2026},
  publisher={Oxford University Press}
}

@article{zhang2025multimodal,
  title={A Multimodal Drug-Target Affinity Prediction Framework with Pretrained Models and Hierarchical Graph Transformer},
  author={Zhang, Zhijun and Liu, Yuanhao and Qu, Jia and Jiao, Zhuqing},
  journal={Journal of Chemical Information and Modeling},
  volume={66},
  number={1},
  pages={310--322},
  year={2025},
  publisher={ACS Publications}
}

@article{han2023himgnn,
  title={HimGNN: a novel hierarchical molecular graph representation learning framework for property prediction},
  author={Han, Shen and Fu, Haitao and Wu, Yuyang and Zhao, Ganglan and Song, Zhenyu and Huang, Feng and Zhang, Zhongfei and Liu, Shichao and Zhang, Wen},
  journal={Briefings in Bioinformatics},
  volume={24},
  number={5},
  pages={bbad305},
  year={2023},
  publisher={Oxford University Press}
}

@article{wojcikowski2015open,
  title={Open Drug Discovery Toolkit (ODDT): a new open-source player in the drug discovery field},
  author={W{\'o}jcikowski, Maciej and Zielenkiewicz, Piotr and Siedlecki, Pawel},
  journal={Journal of cheminformatics},
  volume={7},
  number={1},
  pages={26},
  year={2015},
  publisher={Springer}
}

@article{xu2018powerful,
  title={How powerful are graph neural networks?},
  author={Xu, Keyulu and Hu, Weihua and Leskovec, Jure and Jegelka, Stefanie},
  journal={arXiv preprint arXiv:1810.00826},
  year={2018}
}

@article{liu2022cb,
  title={CB-Dock2: improved protein--ligand blind docking by integrating cavity detection, docking and homologous template fitting},
  author={Liu, Yang and Yang, Xiaocong and Gan, Jianhong and Chen, Shuang and Xiao, Zhi-Xiong and Cao, Yang},
  journal={Nucleic acids research},
  volume={50},
  number={W1},
  pages={W159--W164},
  year={2022},
  publisher={Oxford University Press}
}

@article{sigismund2018emerging,
  title={Emerging functions of the EGFR in cancer},
  author={Sigismund, Sara and Avanzato, Daniele and Lanzetti, Letizia},
  journal={Molecular oncology},
  volume={12},
  number={1},
  pages={3--20},
  year={2018},
  publisher={Wiley Online Library}
}

@article{knox2024drugbank,
  title={DrugBank 6.0: the DrugBank knowledgebase for 2024},
  author={Knox, Craig and Wilson, Mike and Klinger, Christen M and Franklin, Mark and Oler, Eponine and Wilson, Alex and Pon, Allison and Cox, Jordan and Chin, Na Eun and Strawbridge, Seth A and others},
  journal={Nucleic acids research},
  volume={52},
  number={D1},
  pages={D1265--D1275},
  year={2024},
  publisher={Oxford University Press}
}

@article{jumper2021highly,
  title={Highly accurate protein structure prediction with AlphaFold},
  author={Jumper, John and Evans, Richard and Pritzel, Alexander and Green, Tim and Figurnov, Michael and Ronneberger, Olaf and Tunyasuvunakool, Kathryn and Bates, Russ and {\v{Z}}{\'\i}dek, Augustin and Potapenko, Anna and others},
  journal={nature},
  volume={596},
  number={7873},
  pages={583--589},
  year={2021},
  publisher={Nature Publishing Group UK London}
}

@article{uniprot2019uniprot,
  title={UniProt: a worldwide hub of protein knowledge},
  author={UniProt Consortium},
  journal={Nucleic acids research},
  volume={47},
  number={D1},
  pages={D506--D515},
  year={2019},
  publisher={Oxford University Press}
}

@article{landrum2013rdkit,
  title={Rdkit documentation},
  author={Landrum, Greg and others},
  journal={Release},
  volume={1},
  number={1-79},
  pages={4},
  year={2013}
}

@article{zdrazil2024chembl,
  title={The ChEMBL Database in 2023: a drug discovery platform spanning multiple bioactivity data types and time periods},
  author={Zdrazil, Barbara and Felix, Eloy and Hunter, Fiona and Manners, Emma J and Blackshaw, James and Corbett, Sybilla and De Veij, Marleen and Ioannidis, Harris and Lopez, David Mendez and Mosquera, Juan F and others},
  journal={Nucleic acids research},
  volume={52},
  number={D1},
  pages={D1180--D1192},
  year={2024},
  publisher={Oxford University Press}
}

@article{zhu2023associative,
author  = {Zhu, Zhiqin and Yao, Zheng and Qi, Guanqiu and Mazur, Neal and Yang, Pan and Cong, Baisen},
title   = {Associative Learning Mechanism for Drug--Target Interaction Prediction},
journal = {CAAI Transactions on Intelligence Technology},
year    = {2023},
volume  = {8},
number  = {4},
pages   = {1558--1577},
doi     = {10.1049/cit2.12194}
}

@article{zhu2023tdgraphdta,
author  = {Zhu, Zhiqin and Yao, Zheng and Zheng, Xin and Qi, Guanqiu and Li, Yuanyuan and Mazur, Neal and Gao, Xinbo and Gong, Yifei and Cong, Baisen},
title   = {Drug--Target Affinity Prediction Method Based on Multi-Scale Information Interaction and Graph Optimization},
journal = {Computers in Biology and Medicine},
year    = {2023},
volume  = {167},
pages   = {107621},
doi     = {10.1016/j.compbiomed.2023.107621}
}

@article{zhu2024mdctdta,
author  = {Zhu, Zhiqin and Zheng, Xin and Qi, Guanqiu and Gong, Yifei and Li, Yuanyuan and Mazur, Neal and Cong, Baisen and Gao, Xinbo},
title   = {Drug--Target Binding Affinity Prediction Model Based on Multi-Scale Diffusion and Interactive Learning},
journal = {Expert Systems with Applications},
year    = {2024},
volume  = {255},
pages   = {124647},
doi     = {10.1016/j.eswa.2024.124647}
}

@article{zhu2025rrgdta,
author  = {Zhu, Zhiqin and Ding, Yan and Qi, Guanqiu and Cong, Baisen and Li, Yuanyuan and Bai, Litao and Gao, Xinbo},
title   = {Drug--Target Affinity Prediction Using Rotary Encoding and Information Retention Mechanisms},
journal = {Engineering Applications of Artificial Intelligence},
year    = {2025},
volume  = {147},
pages   = {110239},
doi     = {10.1016/j.engappai.2025.110239}
}

@article{guo2026circrna,
  author  = {Guo, Yanbu and Zhu, Haokun and Cao, Jinde and Chen, Rui and Zhou, Dongming},
  title   = {Learning Heterogeneous Biological Interactions via Meta-Relation-Guided Dual-Channel Graph Transformer for Circular Ribonucleic Acid Function Prediction},
  journal = {Knowledge-Based Systems},
  year    = {2026},
  volume  = {342},
  pages   = {115929},
  doi     = {10.1016/j.knosys.2026.115929}
}

@article{liu2026diffad,
  author  = {Liu, Qingbo and Xia, Yuanling and Guo, Yanbu and Li, Weihua},
  title   = {{DiffAD}: Diffusion Augmentation for Influenza Virus Antigenic Distance Prediction},
  journal = {Expert Systems with Applications},
  year    = {2026},
  volume  = {298},
  pages   = {129925},
  doi     = {10.1016/j.eswa.2025.129925}
}




  
  

\newpage
\section{For Table of Contents use only}

\begin{figure*}[htbp]
    \centering
    \includegraphics[width=\textwidth]{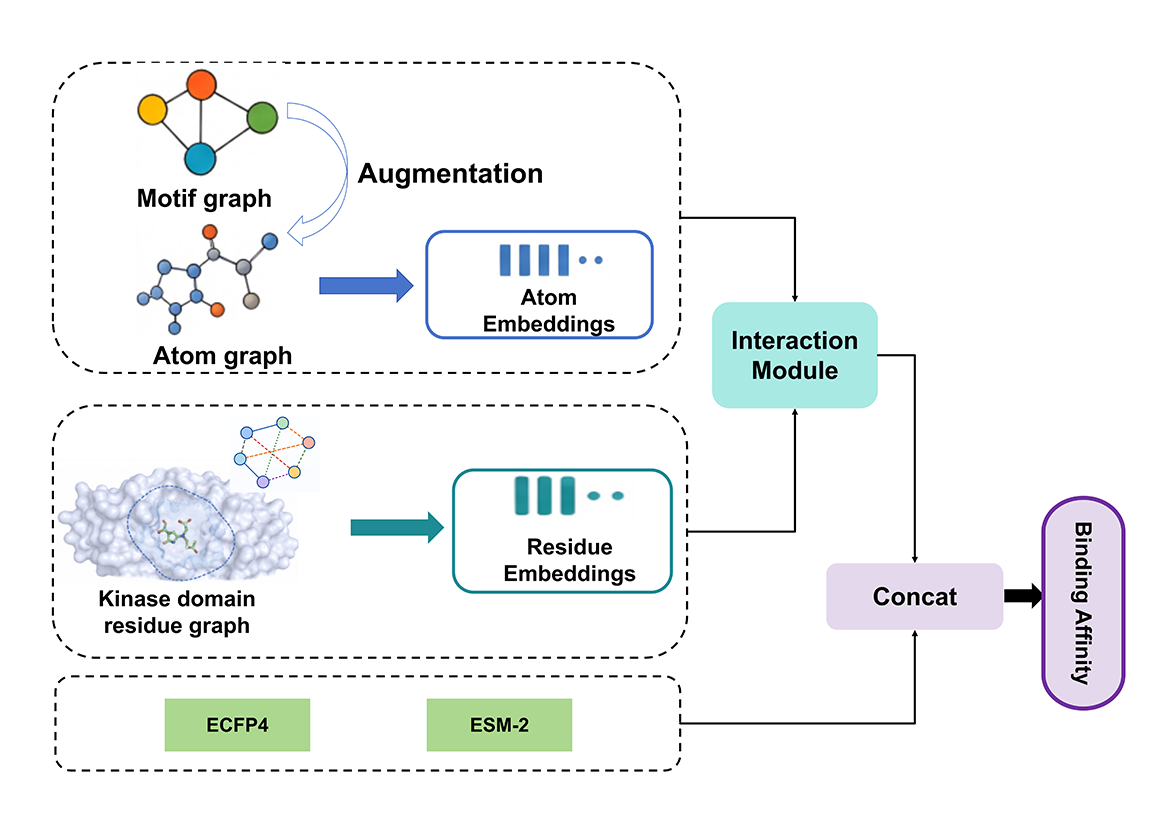}
    \caption{Table of Contents}

\end{figure*}

\end{document}